\documentclass{article}

\usepackage[preprint]{neurips_2025}

\usepackage[utf8]{inputenc} 
\usepackage[T1]{fontenc}    
\usepackage{hyperref}       
\usepackage{url}            
\usepackage{booktabs}       
\usepackage{amsfonts}       
\usepackage{nicefrac}       
\usepackage{microtype}      
\usepackage{xcolor}         
\usepackage{amsmath}        
\usepackage{multirow}
\usepackage{multicol}
\usepackage{graphicx}
\usepackage{floatflt}
\usepackage{enumitem}
\usepackage{subcaption}

\usepackage{xcolor}
\usepackage{tikz}

\usepackage{wrapfig}

\title{Conflict Adaptation in Vision-Language Models}

\author{
Xiaoyang Hu\\[1pt]
Brown University\\
\texttt{xiaoyang\_hu@brown.edu}
}

\begin{document}

\maketitle

\begin{abstract}
A signature of human cognitive control is conflict adaptation: improved performance on a high-conflict trial following another high-conflict trial.
This phenomenon offers an account for how cognitive control, a scarce resource, is recruited.
Using a sequential Stroop task, we find that 12 of 13 vision-language models (VLMs) tested exhibit behavior consistent with conflict adaptation, with the lone exception likely reflecting a ceiling effect.
To understand the representational basis of this behavior, we use sparse autoencoders (SAEs) to identify task-relevant supernodes in InternVL 3.5 4B.
Partially overlapping supernodes emerge for text and color in both early and late layers, and their relative sizes mirror the automaticity asymmetry between reading and color naming in humans.
We further isolate a conflict-modulated supernode in layers 24–25 whose ablation significantly increases Stroop errors while minimally affecting congruent trials.
\end{abstract}

\begin{figure}[htbp]
    \centering
    \setlength{\fboxrule}{0.6pt}
    \hspace{25pt}
    \begin{subfigure}{0.4\textwidth}
        \centering
        \fbox{\includegraphics[width=\dimexpr\textwidth-2\fboxsep-2\fboxrule]{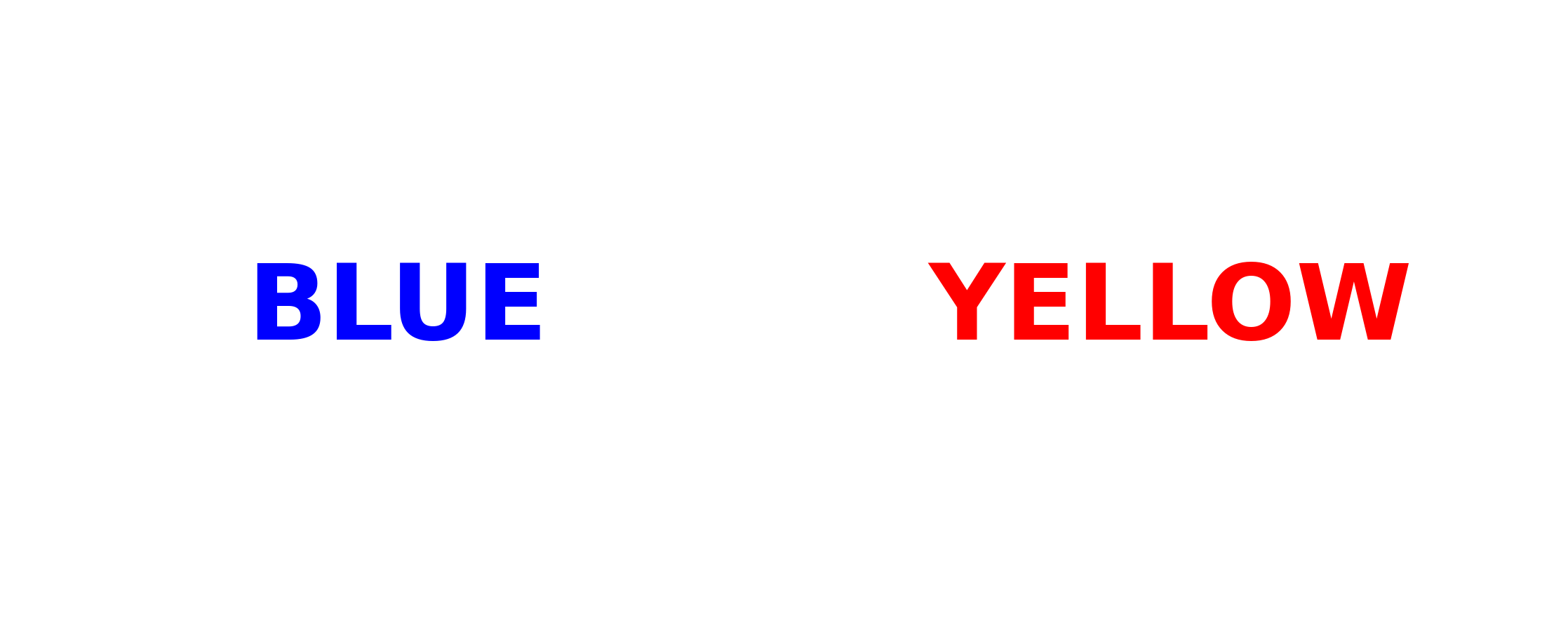}}
        \label{fig:1a}
    \end{subfigure}
    \hfill
    \begin{subfigure}{0.4\textwidth}
        \centering
        \fbox{\includegraphics[width=\dimexpr\textwidth-2\fboxsep-2\fboxrule]{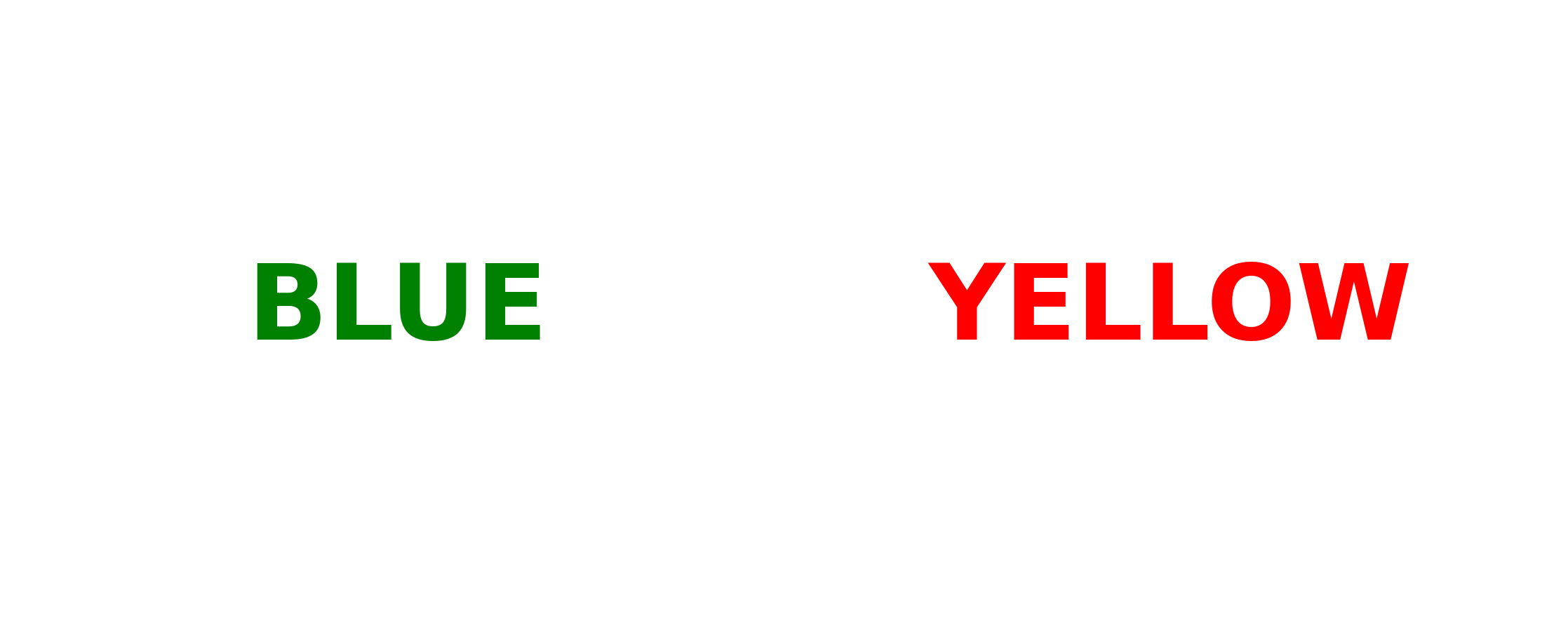}}
        \label{fig:1b}
    \end{subfigure}
    \hspace{25pt}
    \vspace{-5pt}
    \caption{Sequential Stroop task design. Left: a congruent trial followed by an incongruent trial. Right: an incongruent trial followed by another incongruent trial.}
    \label{fig:task}
\end{figure}

\section{Introduction}\label{sec:intro}

Inhibitory control—the ability to suppress prepotent responses in favor of goal-appropriate ones—is fundamental to human cognition. The Stroop task, where participants must name the color of text while ignoring the word itself, remains the canonical paradigm for studying inhibitory control~\cite{stroop1935studies}. Recent work has begun investigating how VLMs handle conflicting information from different modalities: Luo et al.~\cite{luo2025machine} applied Stroop and flanker tasks to VLMs, demonstrating robust congruency effects across a large number of models, while Hua et al.~\cite{hua2025vision} and Ortu et al.~\cite{ortu2025seeing} examined visual-textual conflicts in VLMs.

A signature of human cognitive control is conflict adaptation: improved performance on incongruent trials when they follow other incongruent trials compared to congruent trials.
This effect offers an account for how cognitive control, a scarce resource, is recruited~\cite{botvinick2001conflict,gratton1992optimizing}.
Whether VLMs exhibit conflict adaptation, and through what mechanisms, is poorly understood.
Understanding the emergence of such adaptation would inform theories of goal-directed processing across artificial and natural minds.
\section{Experiments}\label{sec:results}

We implemented a sequential Stroop task where models are presented with images containing two color words in colored fonts, positioned either left-to-right or top-to-bottom. Each word's font color may match (congruent) or mismatch (incongruent) its semantic meaning (Figure~\ref{fig:task}). Crucially, the two words do not share any colors in either modality.
Our color set includes red, blue, green, yellow, pink, and brown, extending beyond the traditional red-blue-green Stroop paradigm.
This design yields 30 CC (congruent-congruent) sequences, 120 CI (congruent-incongruent) sequences, 120 IC (incongruent-congruent) sequences, and 360 II (incongruent-incongruent) sequences per arrangement.

Models were prompted to name the ink colors in exactly two words.
They received as system prompts explicit task instructions designed to minimize ambiguity:
\textit{``You are a participant in a cognitive task. You will see an image with two words positioned from }\{\textit{left to right / top to bottom}\}\textit{. Your task is to name the color of the ink each word is printed in. Do not read what the words say. Only report the actual ink colors. Answer in exactly two words: first the }\{\textit{left / top}\}\textit{ ink color, then the }\{\textit{right / bottom}\}\textit{ ink color.''}
We tested 13 leading open-source VLMs from Gemma~\cite{team2025gemma}, InternVL~\cite{wang2025internvl3}, Molmo~\cite{deitke2025molmo}, and Qwen~\cite{bai2025qwen2} families.
Each model was evaluated on all sequences for both spatial arrangements, with performance measured by the log probabilities assigned to correct second color tokens.\footnote{Models without system prompt support received abbreviated instructions after the image but nevertheless performed well due to strong baseline capabilities.}

\subsection{Behavioral Results}

\begin{figure}[htbp]
    \centering
    \begin{subfigure}[b]{0.161\textwidth}
        \centering
        \includegraphics[width=\textwidth]{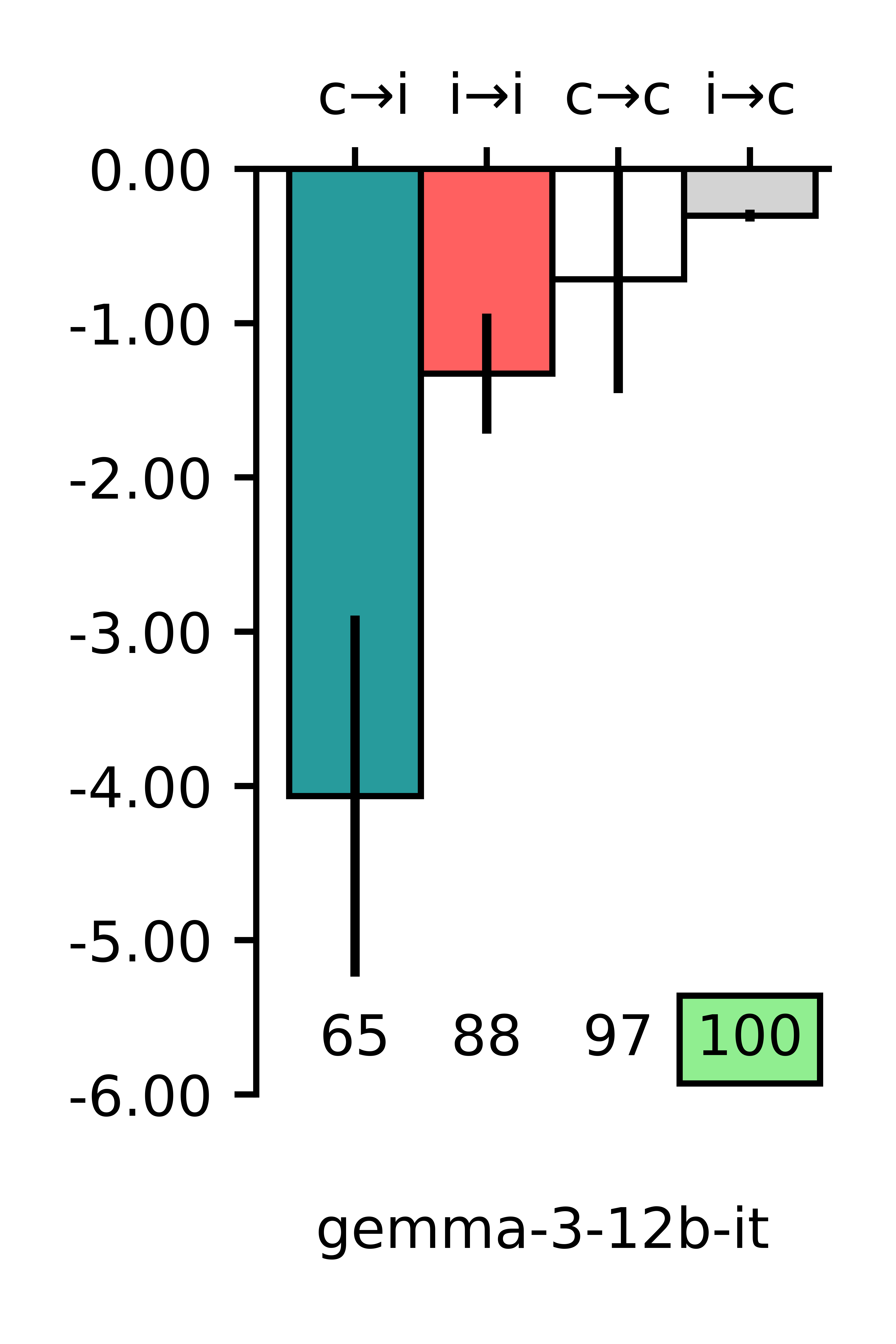}
    \end{subfigure}
    \begin{subfigure}[b]{0.161\textwidth}
        \centering
        \includegraphics[width=\textwidth]{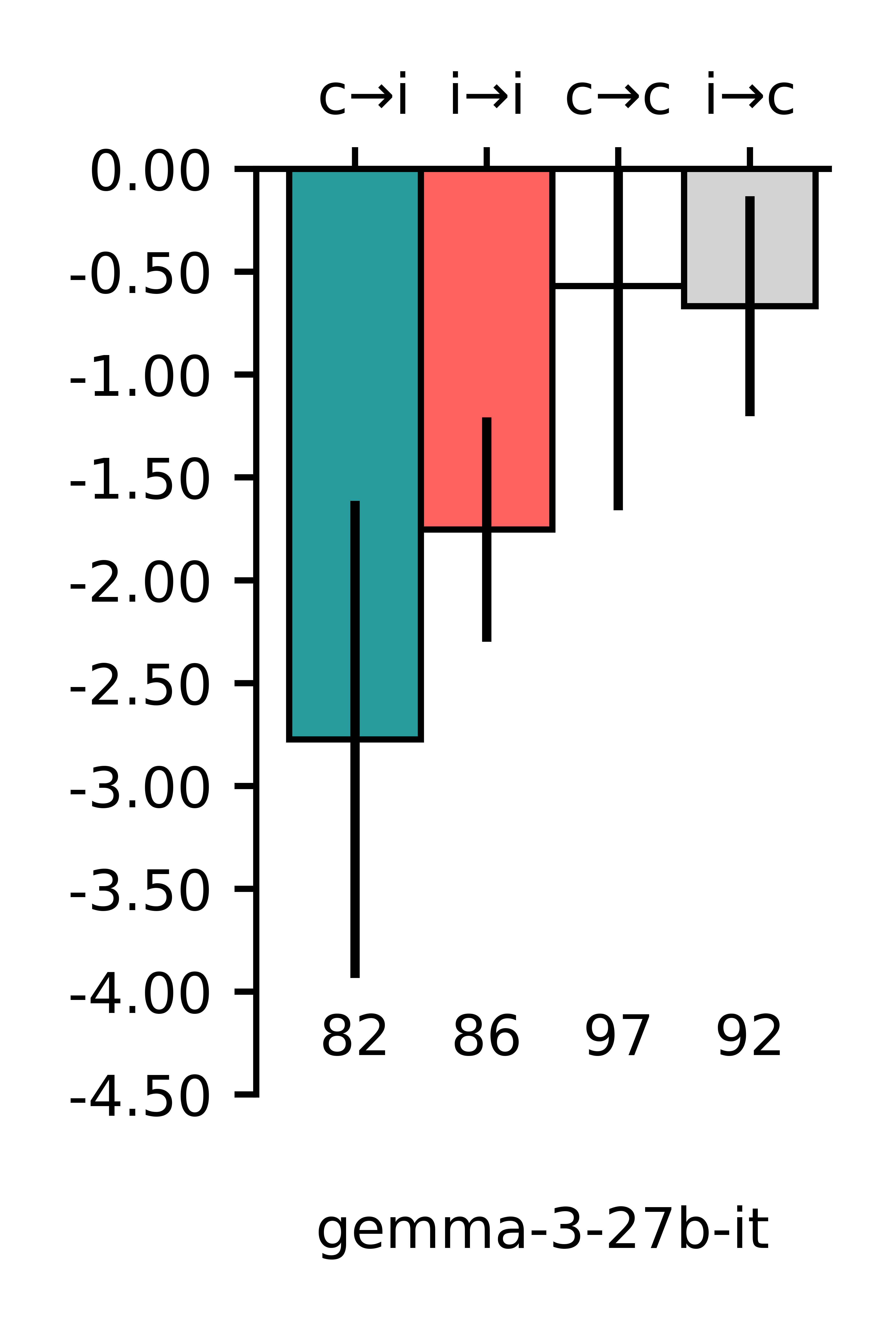}
    \end{subfigure}
    \begin{subfigure}[b]{0.161\textwidth}
        \centering
        \includegraphics[width=\textwidth]{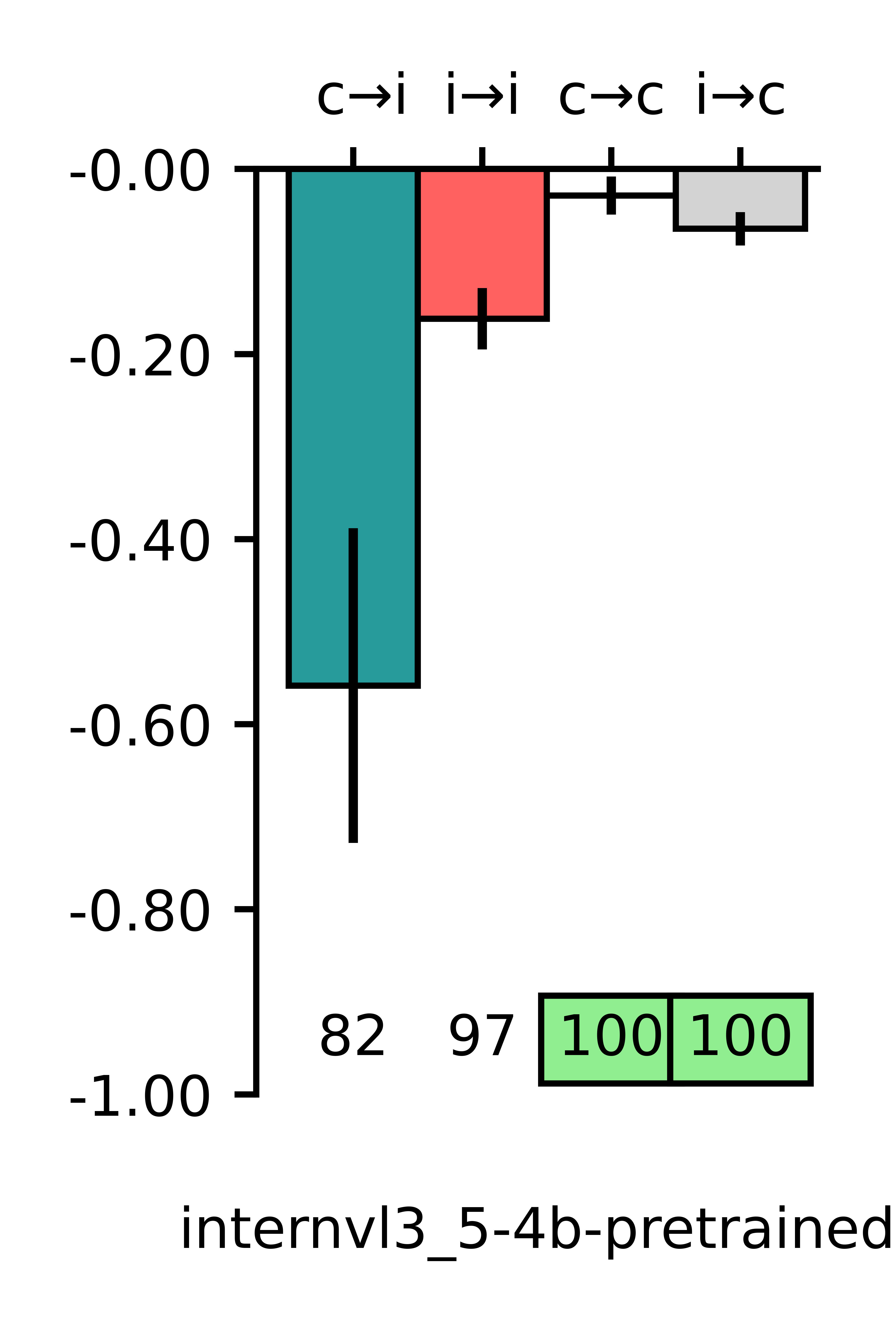}
    \end{subfigure}
    \begin{subfigure}[b]{0.161\textwidth}
        \centering
        \includegraphics[width=\textwidth]{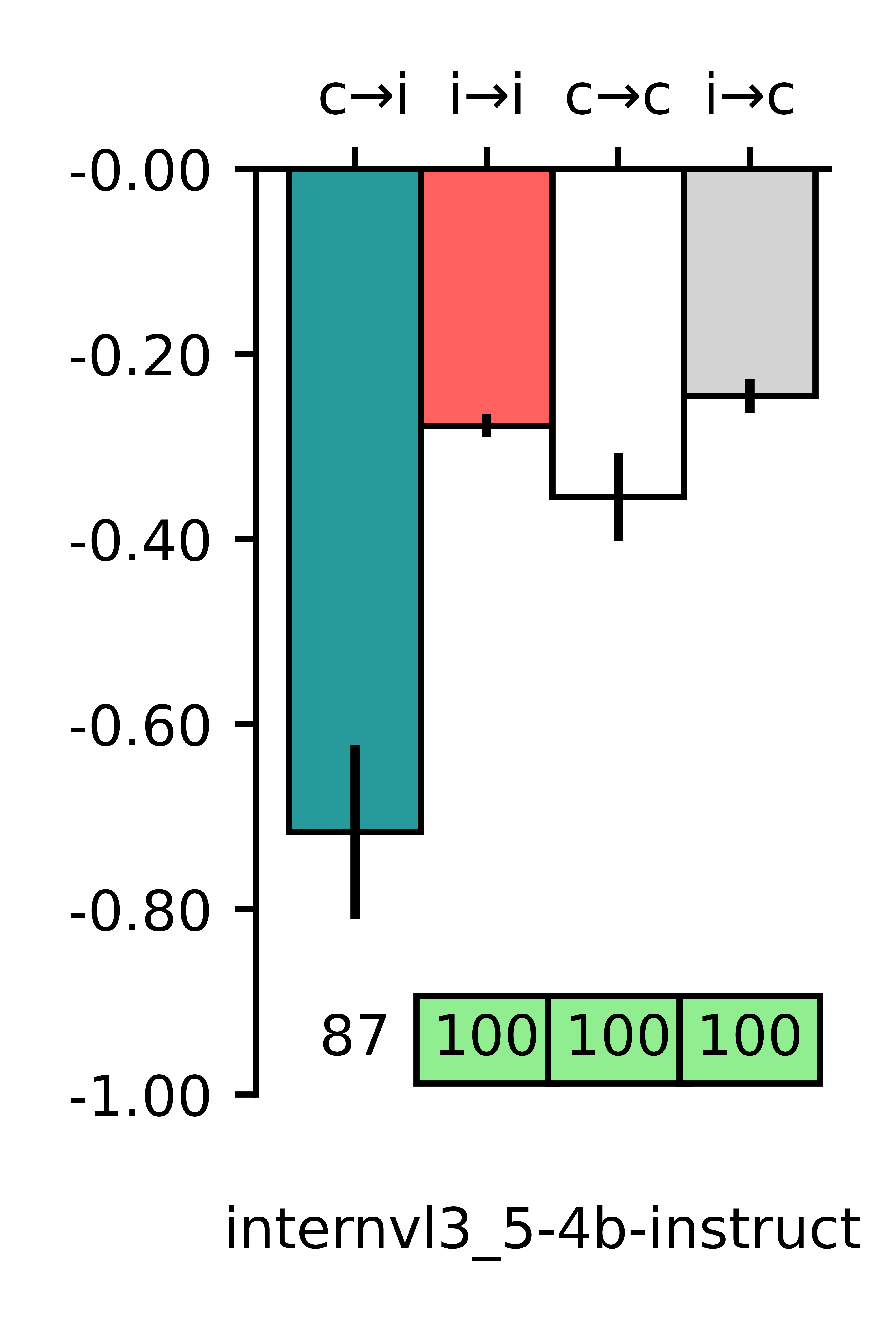}
    \end{subfigure}
    \begin{subfigure}[b]{0.161\textwidth}
        \centering
        \includegraphics[width=\textwidth]{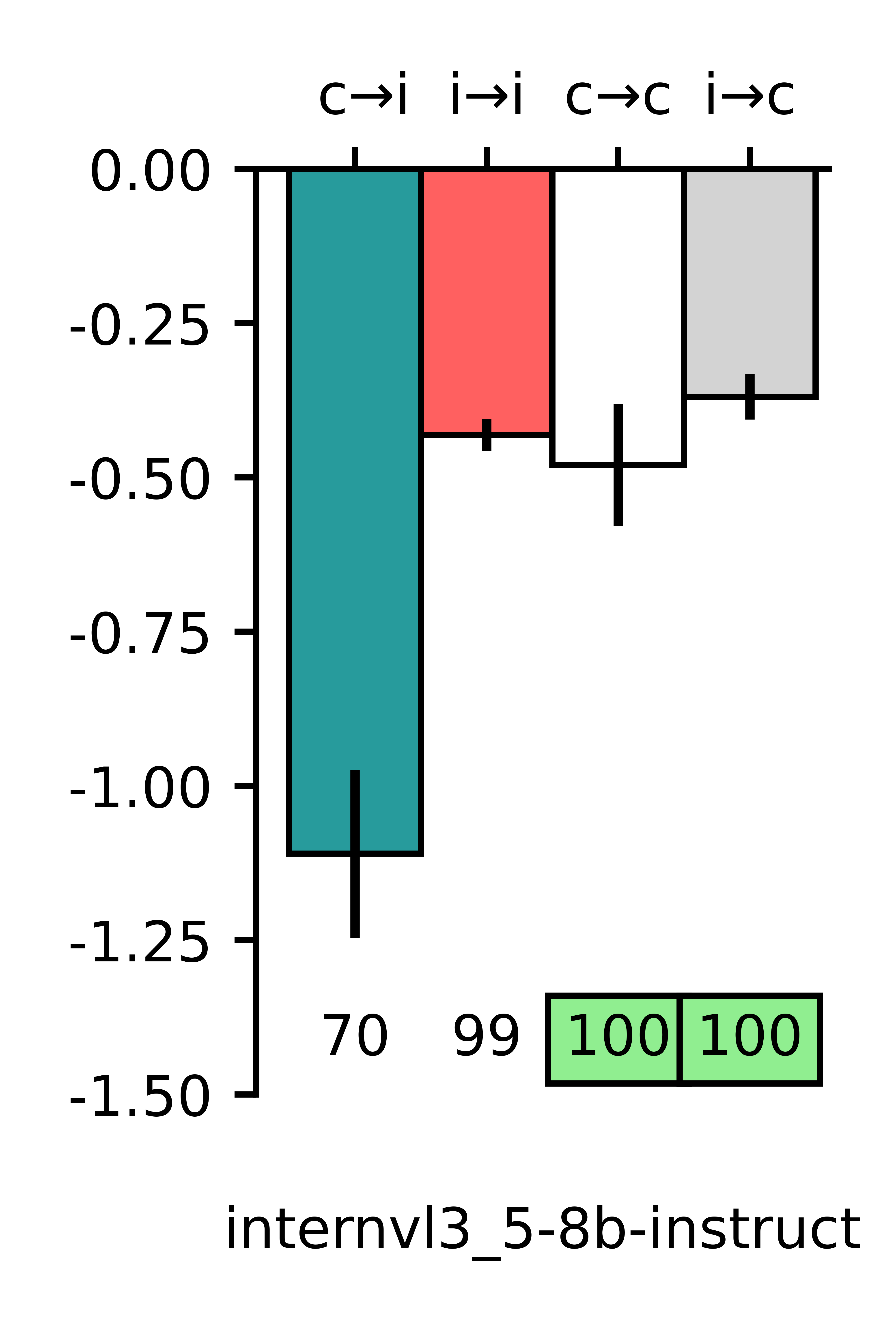}
    \end{subfigure}
    \begin{subfigure}[b]{0.161\textwidth}
        \centering
        \includegraphics[width=\textwidth]{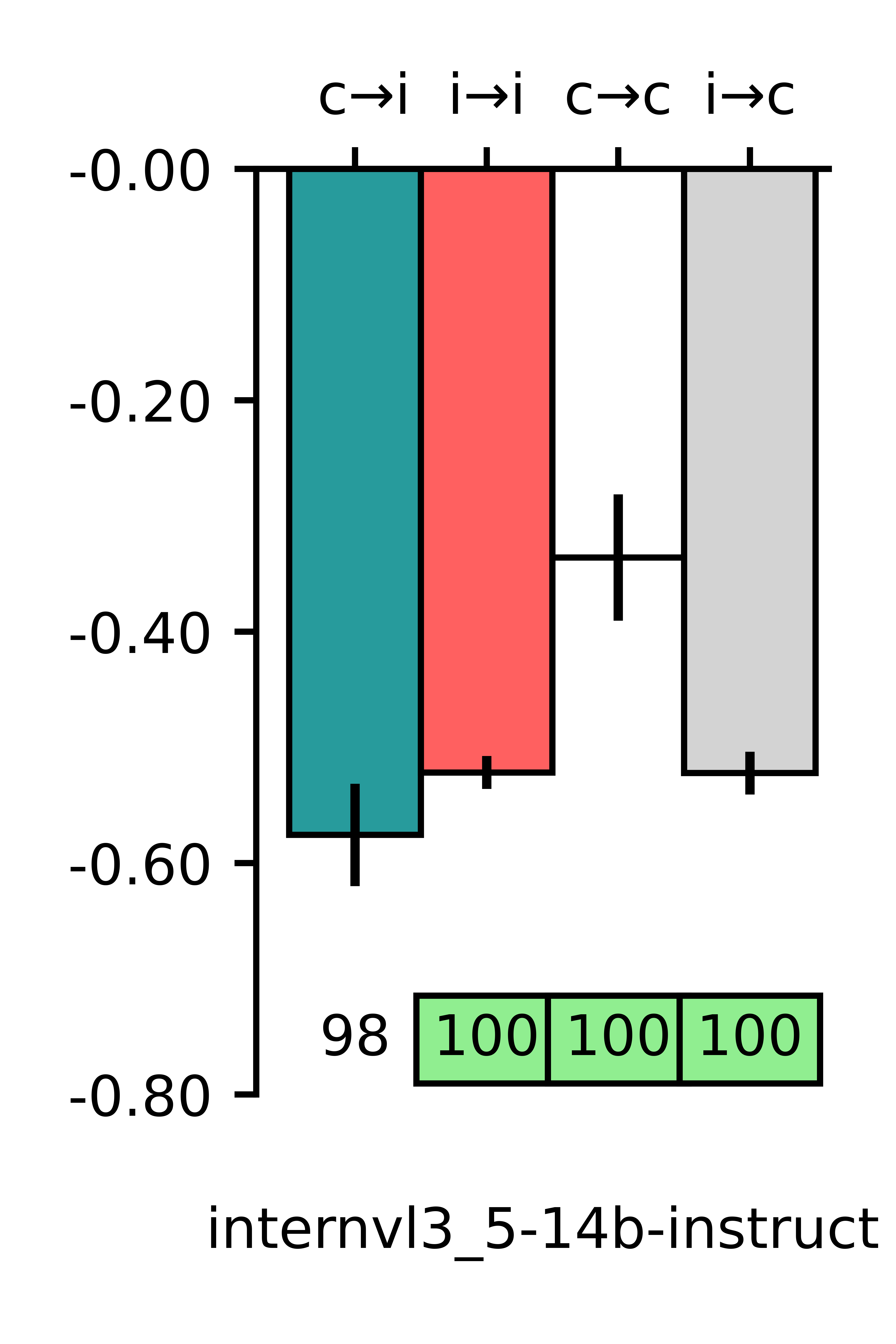}
    \end{subfigure}
    
    \begin{subfigure}[b]{0.161\textwidth}
        \centering
        \includegraphics[width=\textwidth]{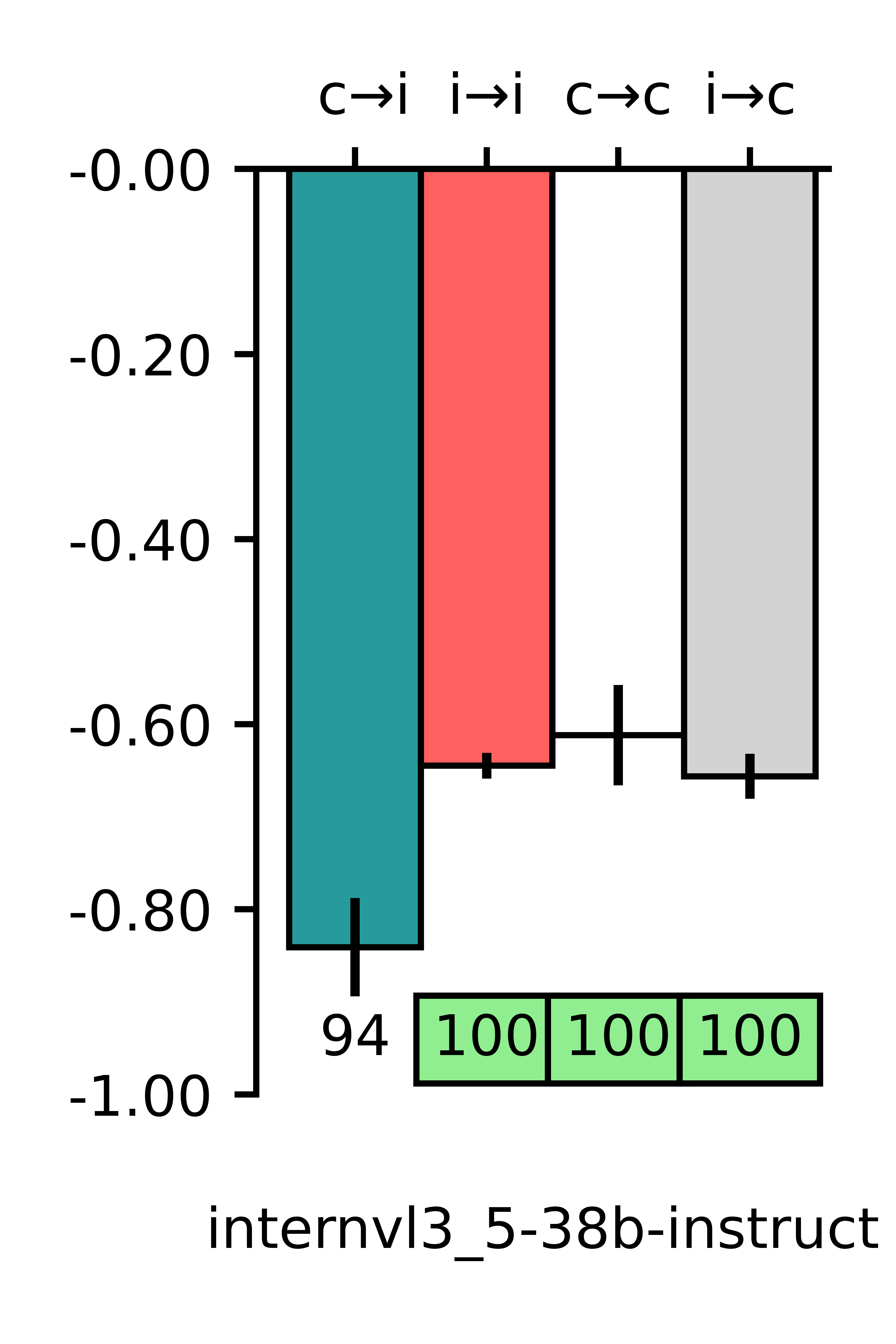}
    \end{subfigure}
    \begin{subfigure}[b]{0.161\textwidth}
        \centering
        \includegraphics[width=\textwidth]{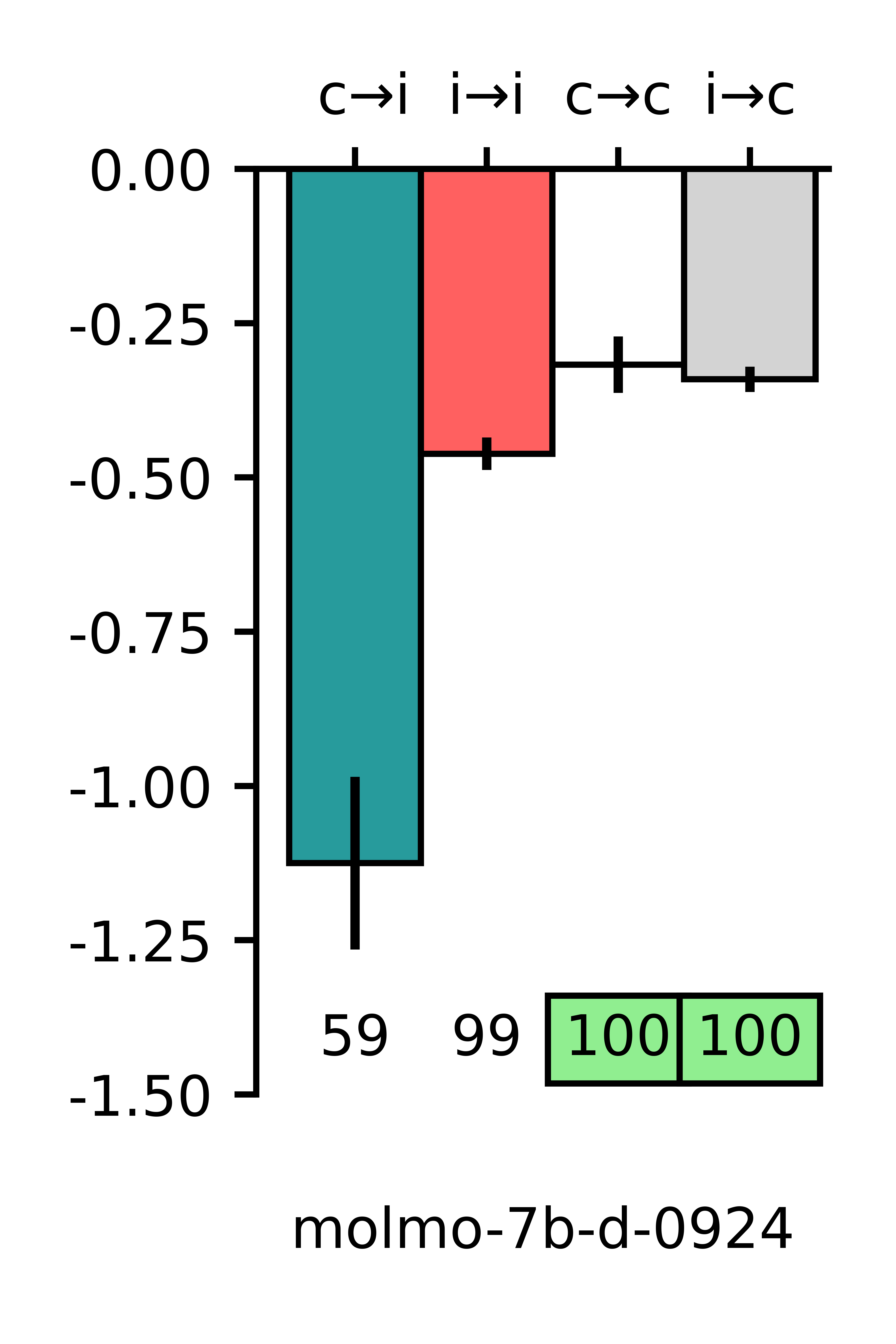}
    \end{subfigure}
    \begin{subfigure}[b]{0.161\textwidth}
        \centering
        \includegraphics[width=\textwidth]{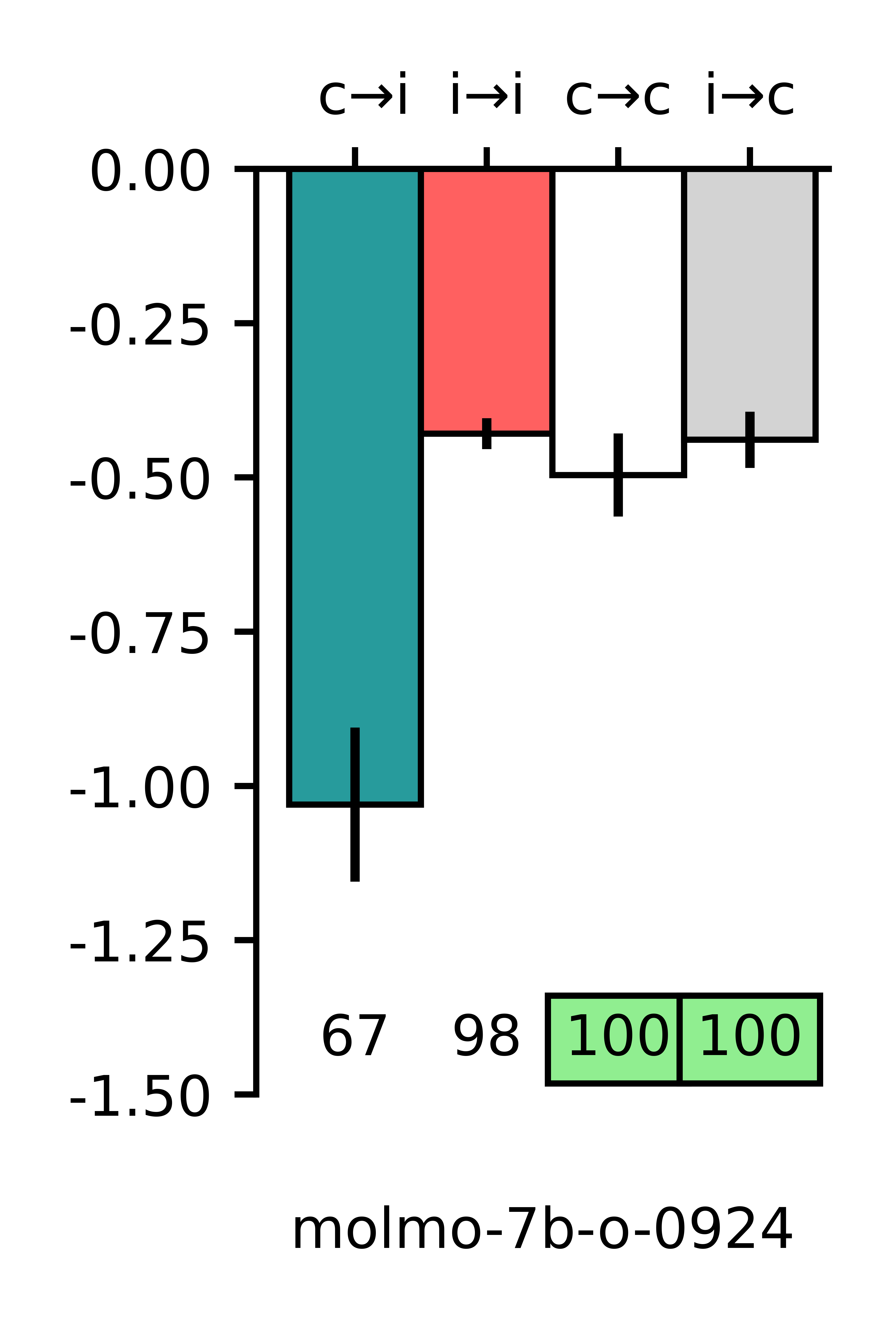}
    \end{subfigure}
    \begin{subfigure}[b]{0.161\textwidth}
        \centering
        \includegraphics[width=\textwidth]{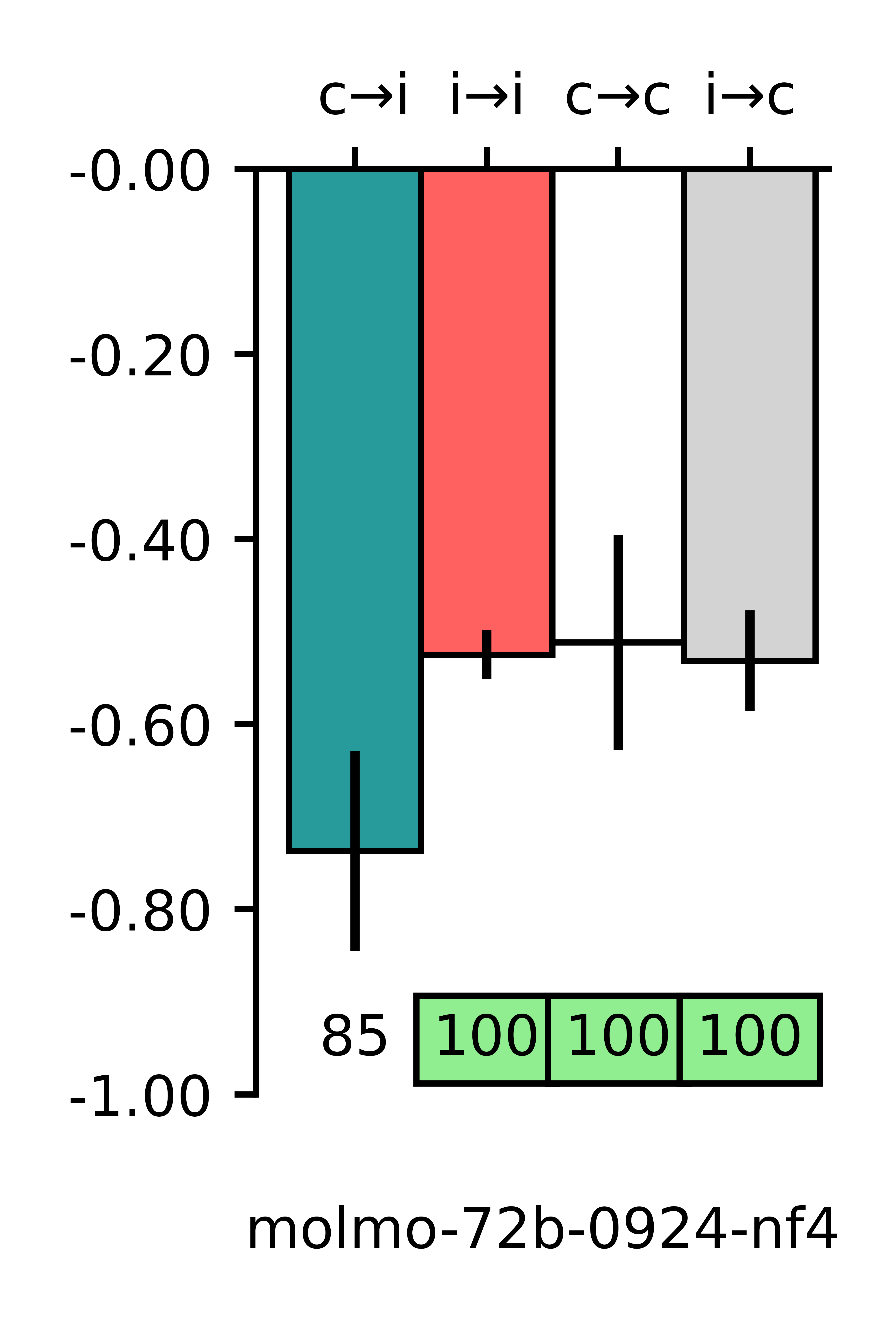}
    \end{subfigure}
    \begin{subfigure}[b]{0.161\textwidth}
        \centering
        \includegraphics[width=\textwidth]{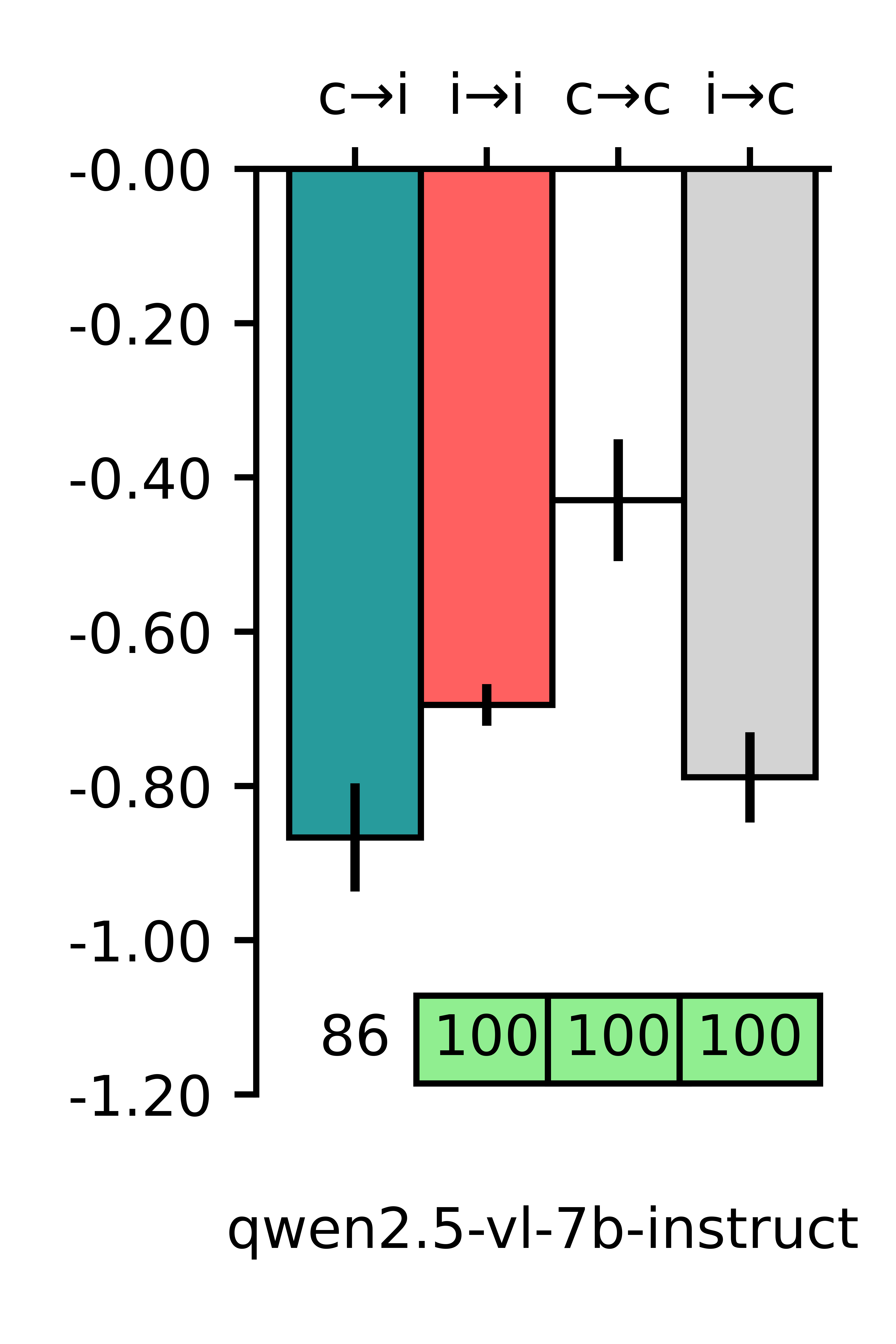}
    \end{subfigure}
    \begin{subfigure}[b]{0.161\textwidth}
        \centering
        \includegraphics[width=\textwidth]{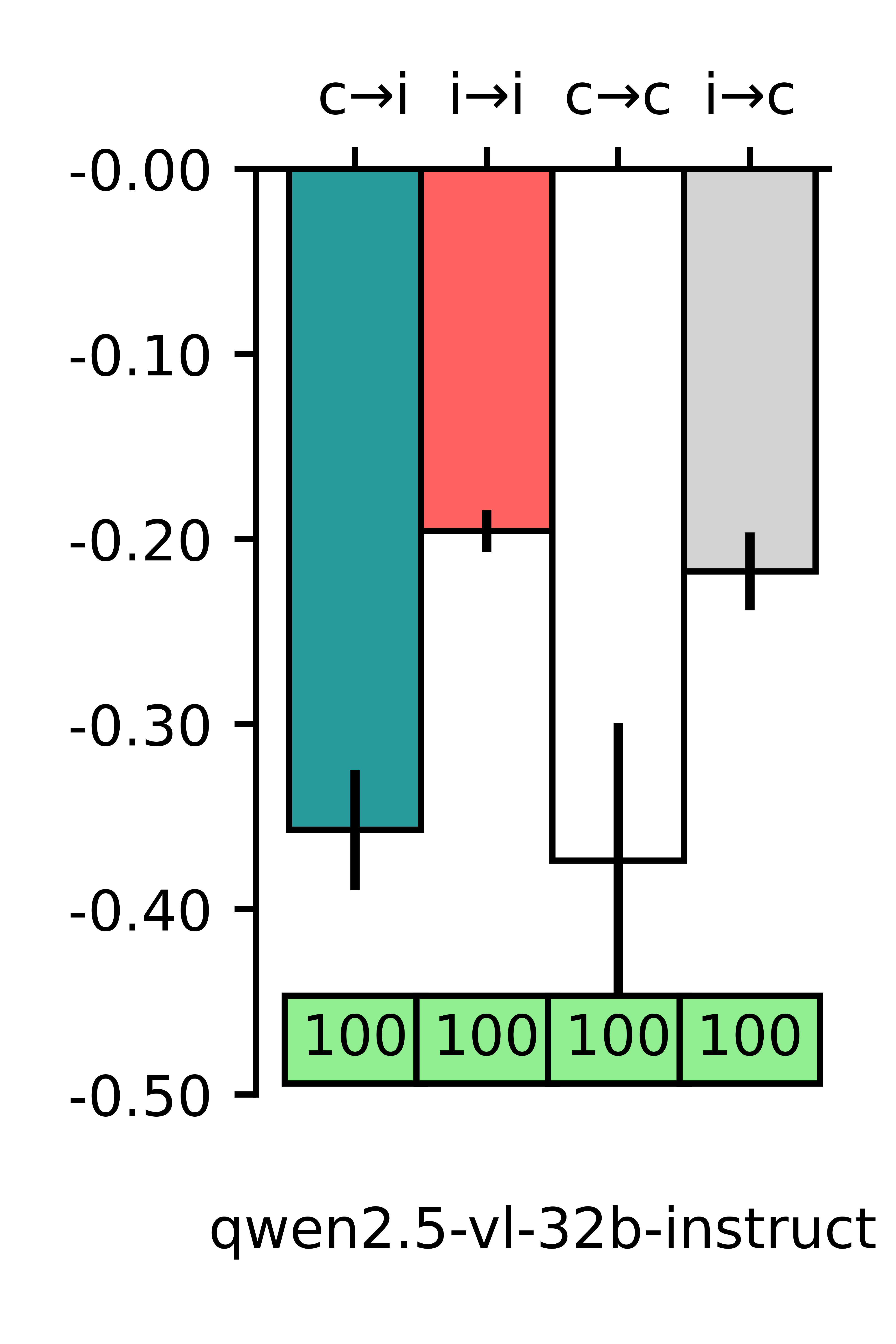}
    \end{subfigure}
    \caption{Average log probabilities assigned to correct second color tokens across conditions under left-right word arrangement. 12 of 13 models tested show higher values for II (incongruent following incongruent) than CI (incongruent following congruent). Condition accuracy shown below each bar.}
    \label{fig:model_logprobs_left_right}
\end{figure}

\begin{figure}[htbp]
    \centering
    \begin{subfigure}[b]{0.161\textwidth}
        \centering
        \includegraphics[width=\textwidth]{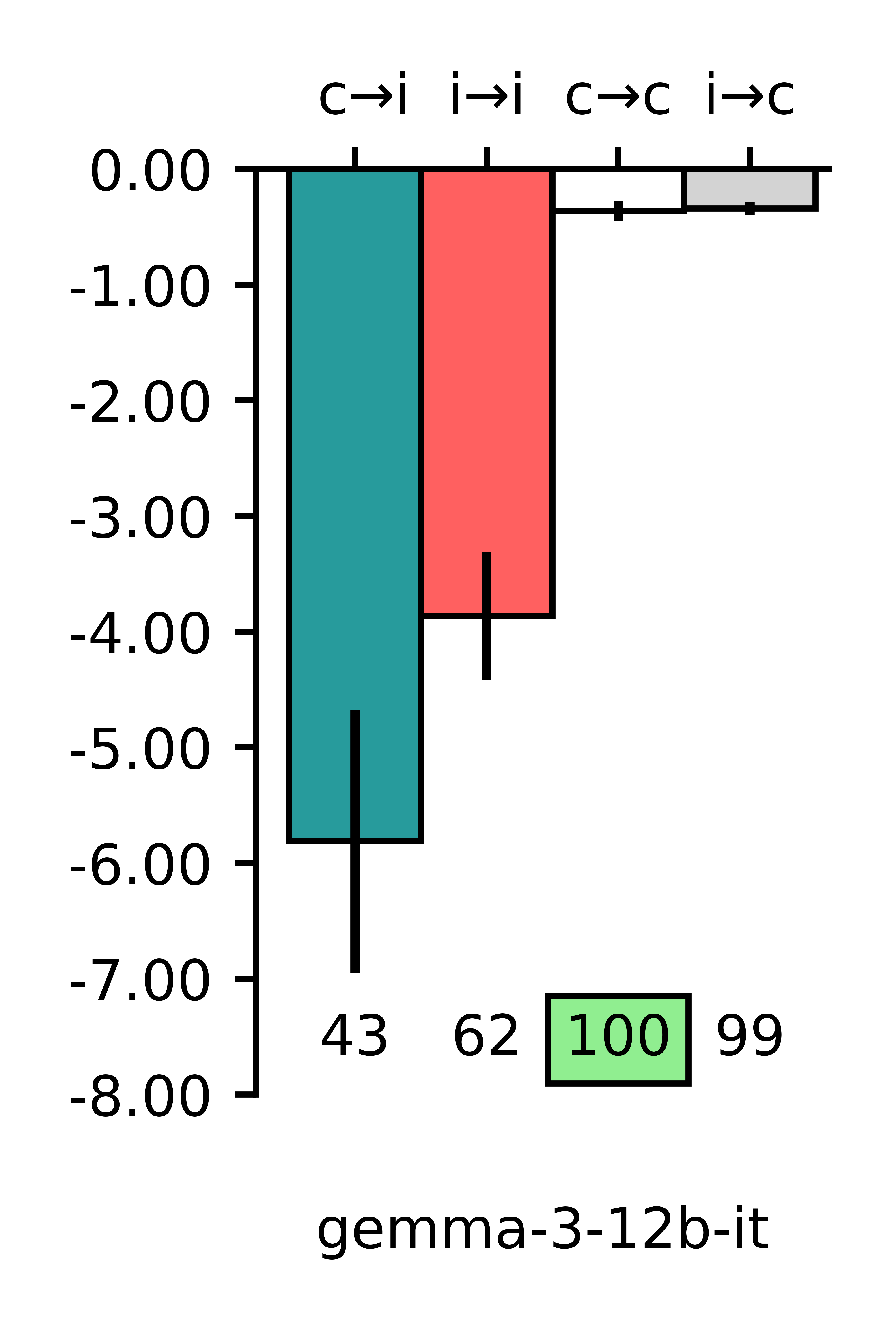}
    \end{subfigure}
    \begin{subfigure}[b]{0.161\textwidth}
        \centering
        \includegraphics[width=\textwidth]{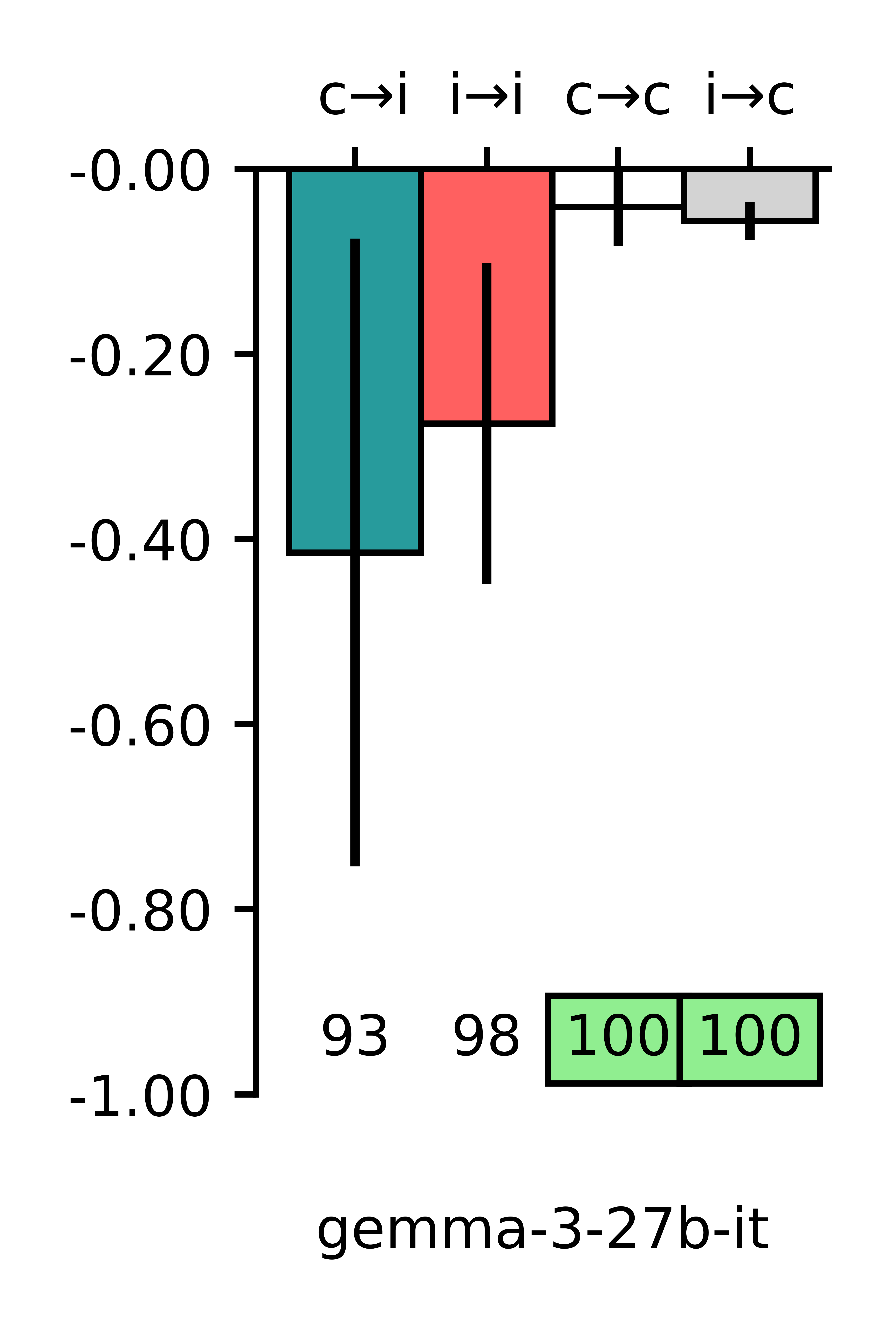}
    \end{subfigure}
    \begin{subfigure}[b]{0.161\textwidth}
        \centering
        \includegraphics[width=\textwidth]{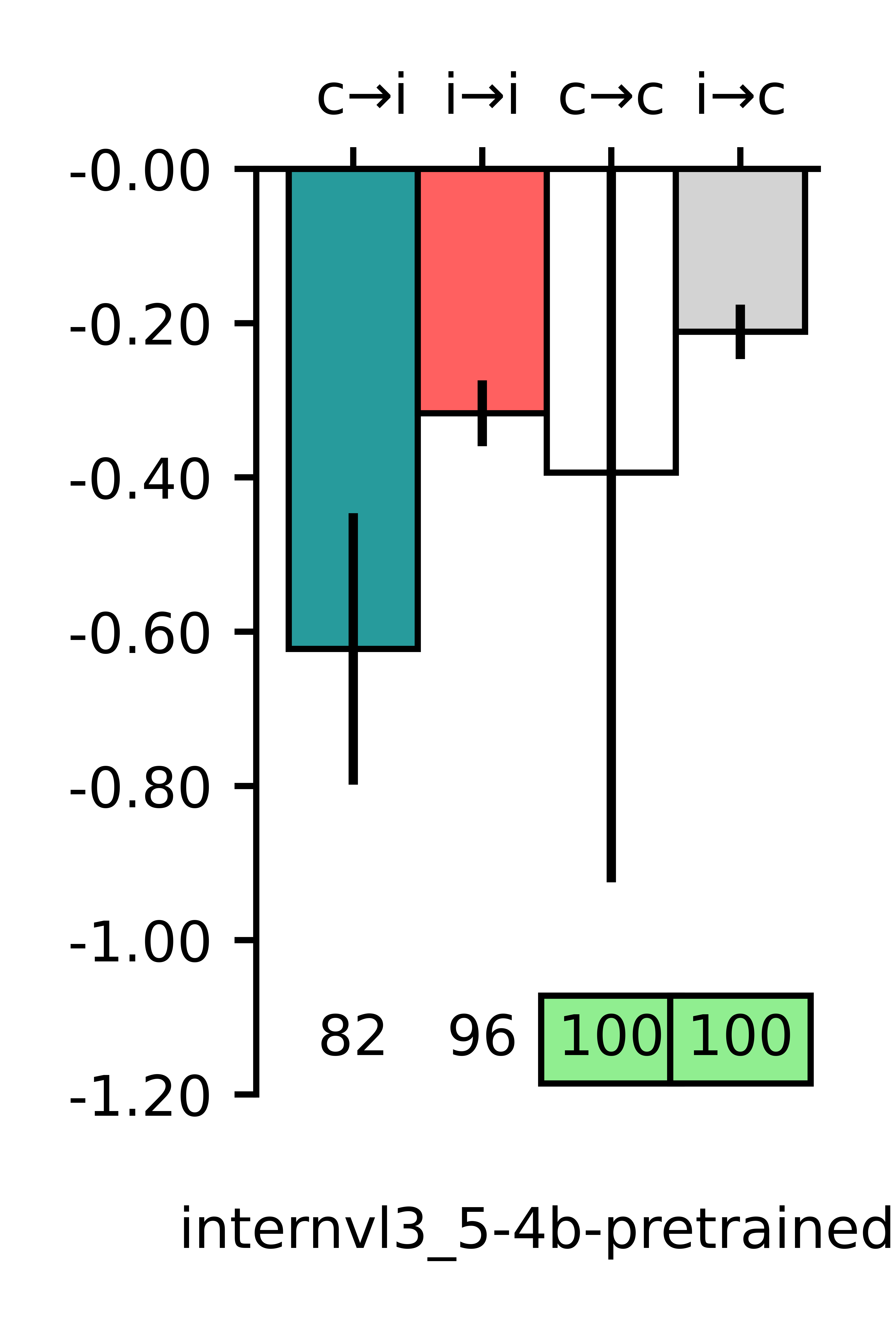}
    \end{subfigure}
    \begin{subfigure}[b]{0.161\textwidth}
        \centering
        \includegraphics[width=\textwidth]{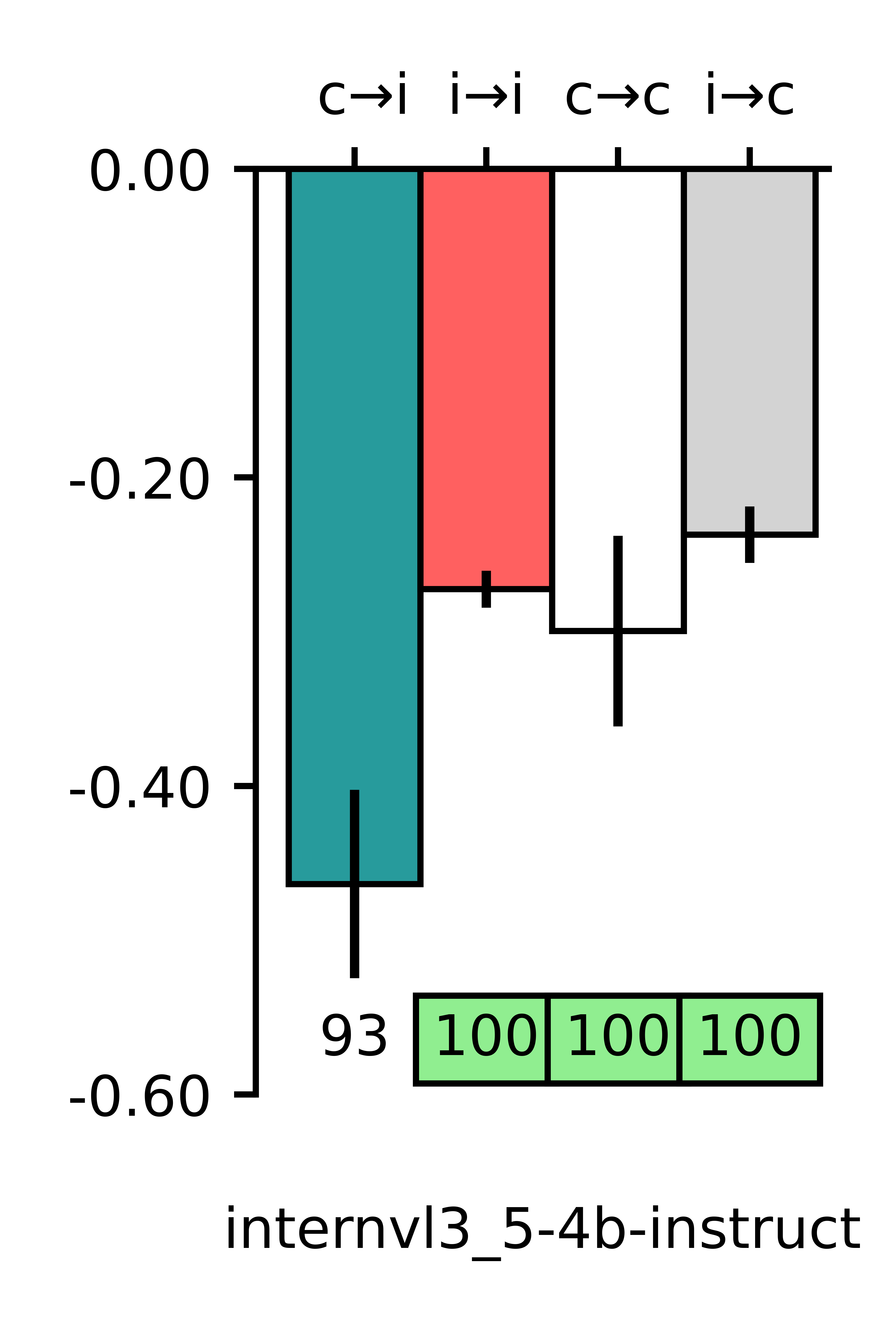}
    \end{subfigure}
    \begin{subfigure}[b]{0.161\textwidth}
        \centering
        \includegraphics[width=\textwidth]{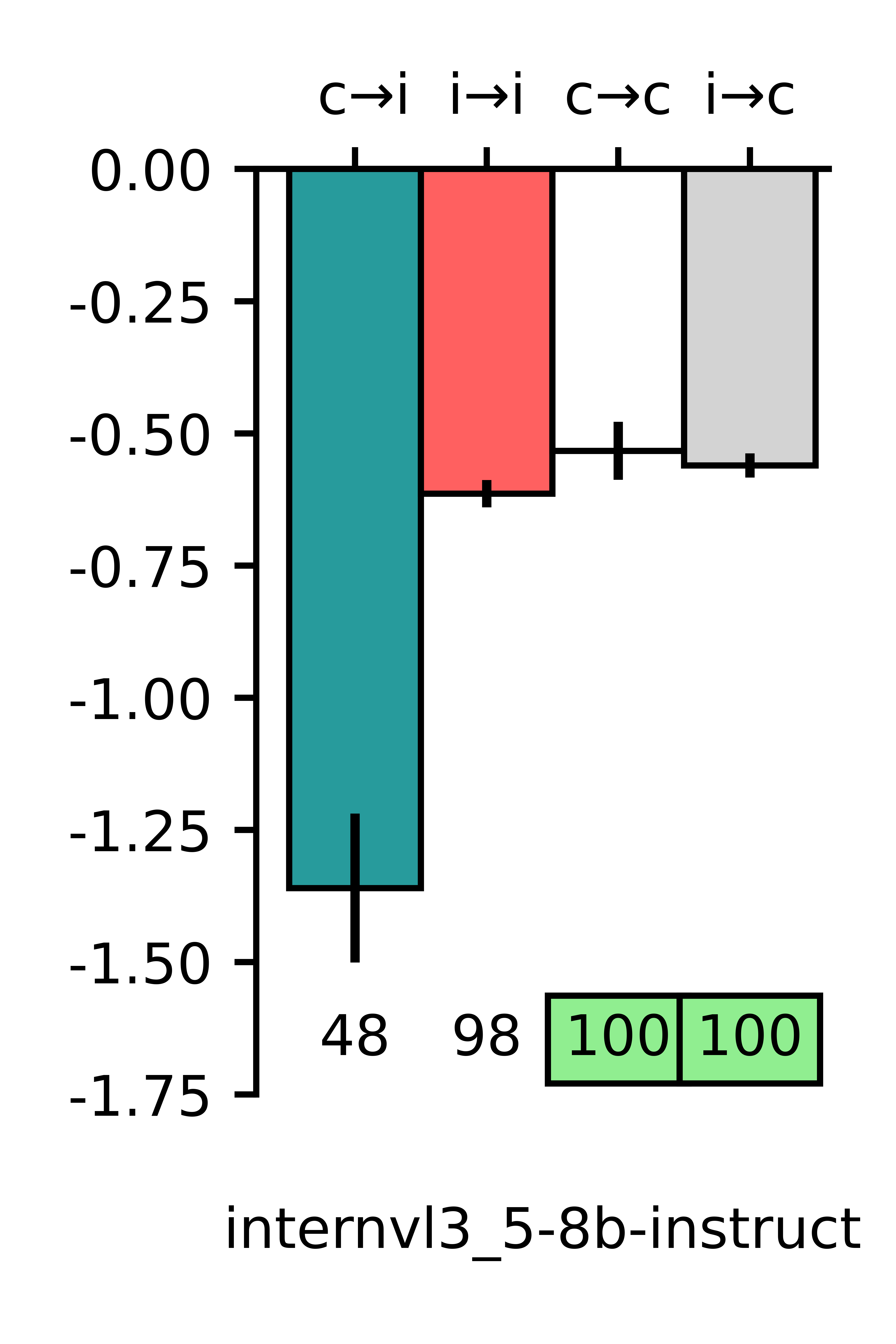}
    \end{subfigure}
    \begin{subfigure}[b]{0.161\textwidth}
        \centering
        \includegraphics[width=\textwidth]{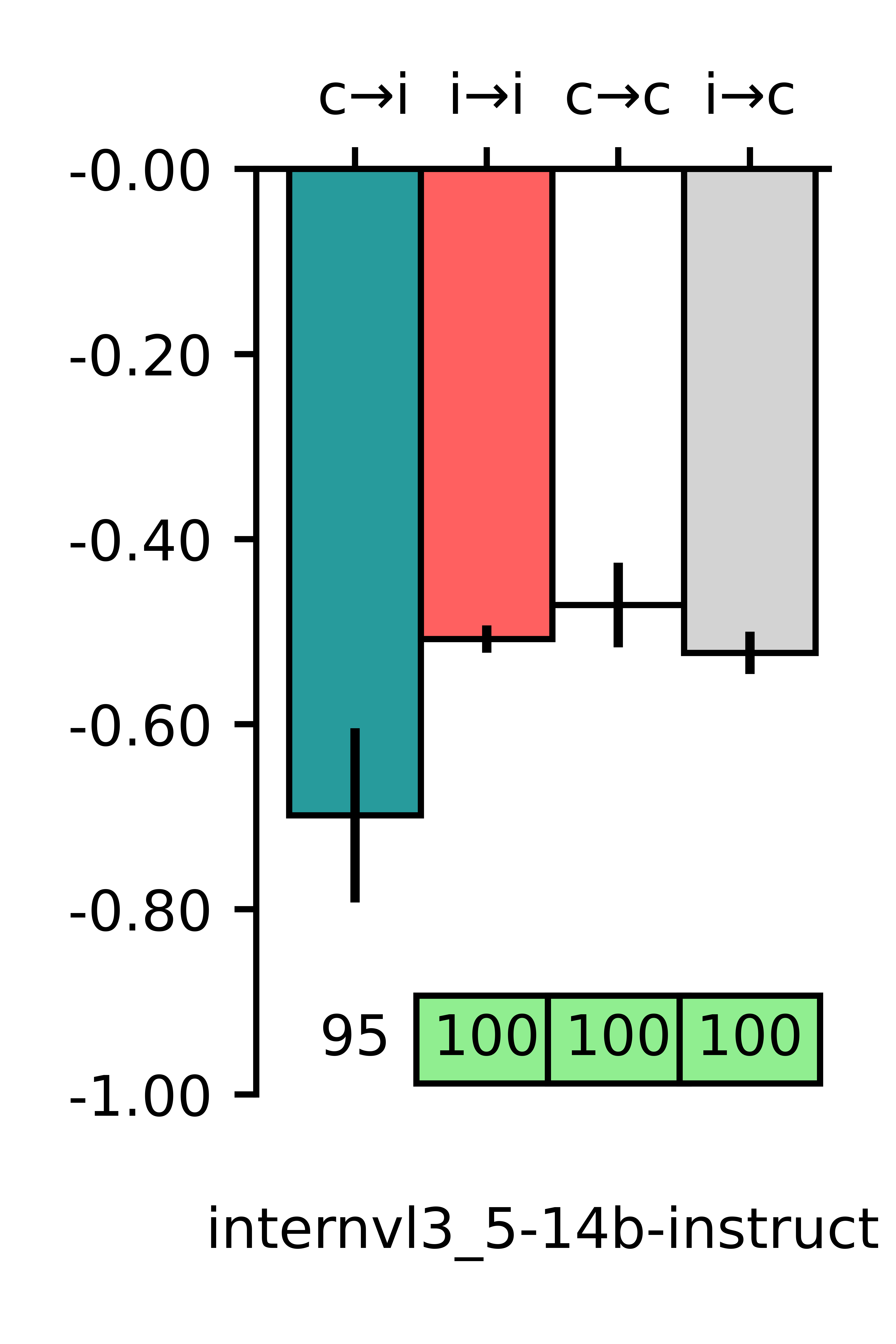}
    \end{subfigure}
    
    \begin{subfigure}[b]{0.161\textwidth}
        \centering
        \includegraphics[width=\textwidth]{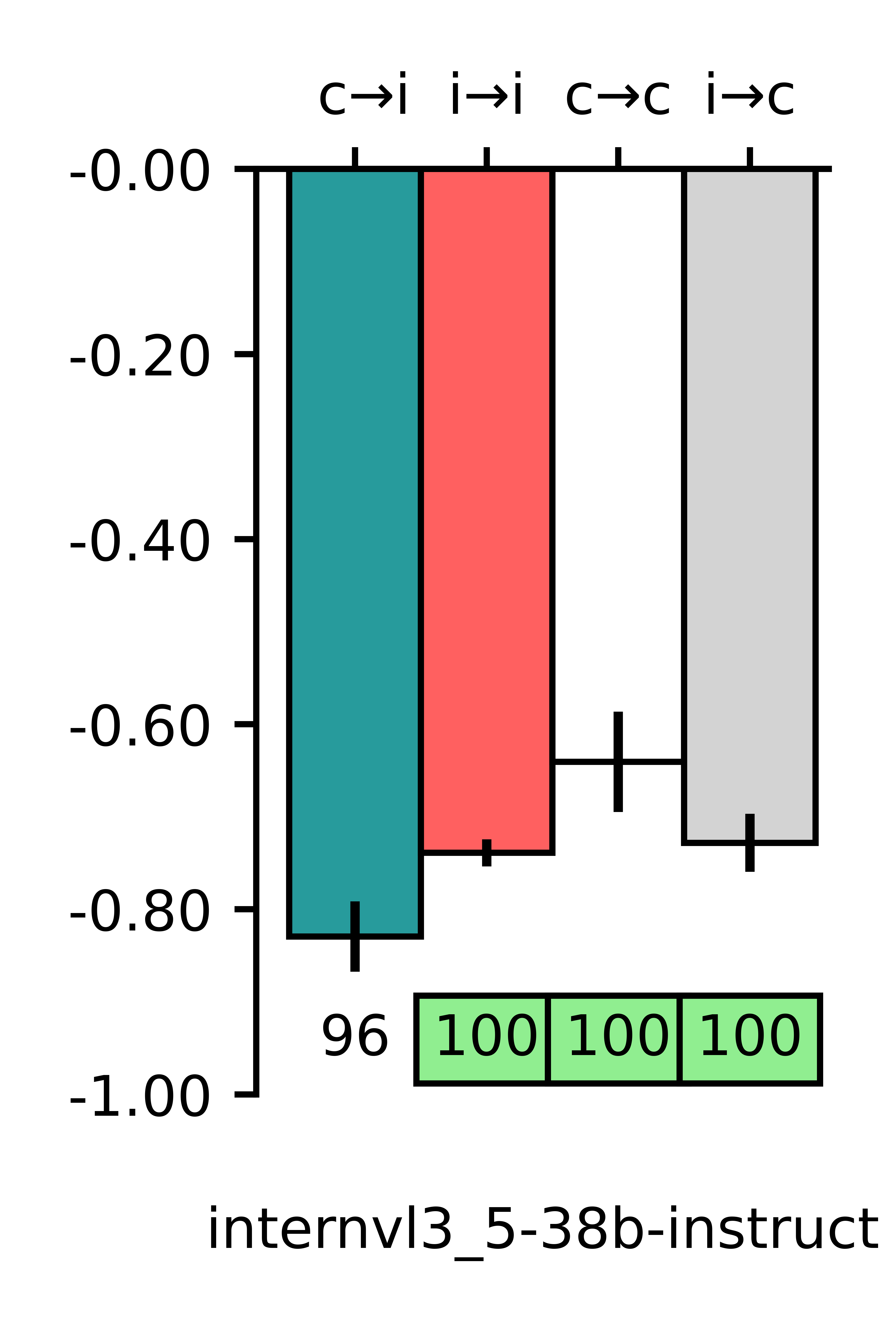}
    \end{subfigure}
    \begin{subfigure}[b]{0.161\textwidth}
        \centering
        \includegraphics[width=\textwidth]{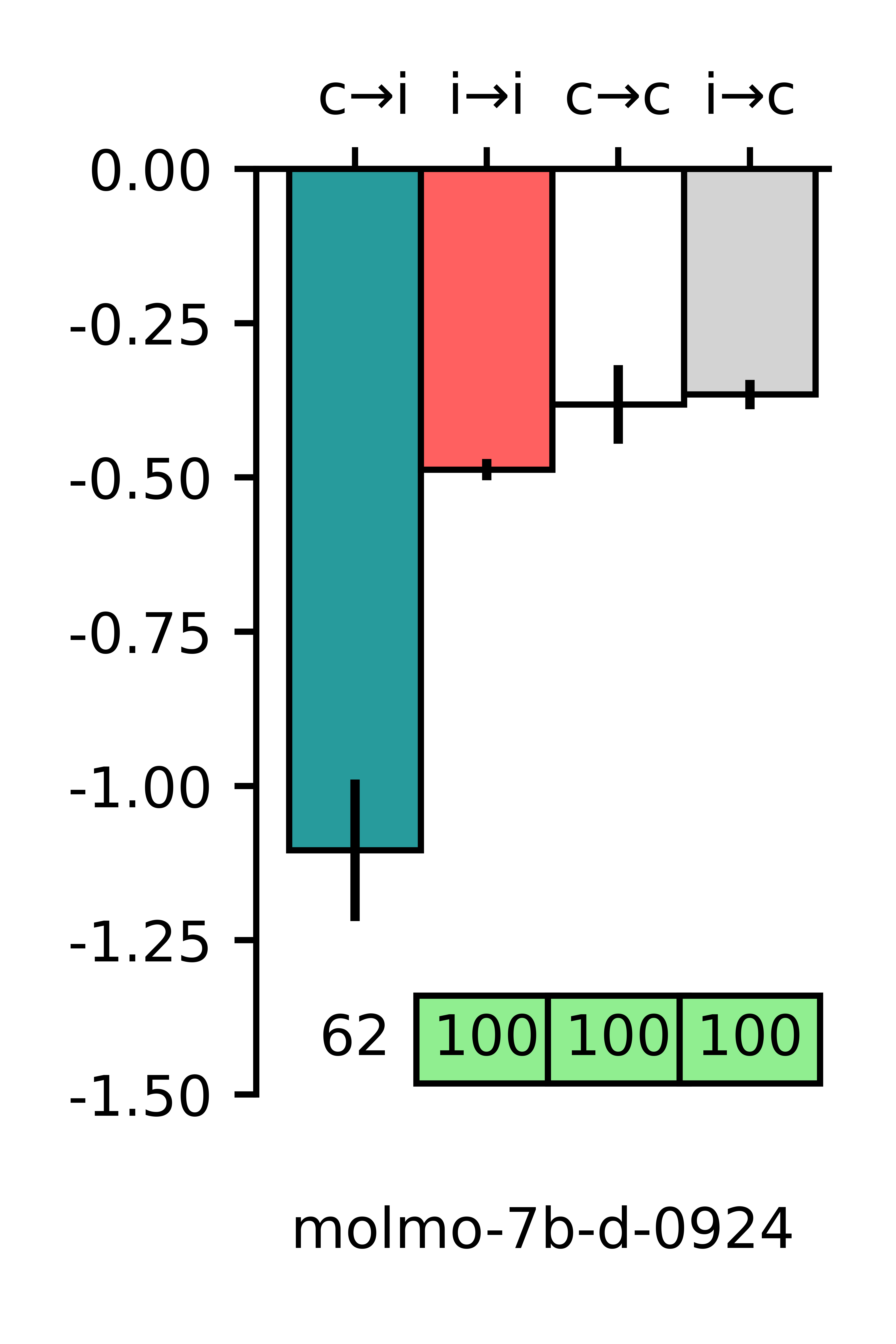}
    \end{subfigure}
    \begin{subfigure}[b]{0.161\textwidth}
        \centering
        \includegraphics[width=\textwidth]{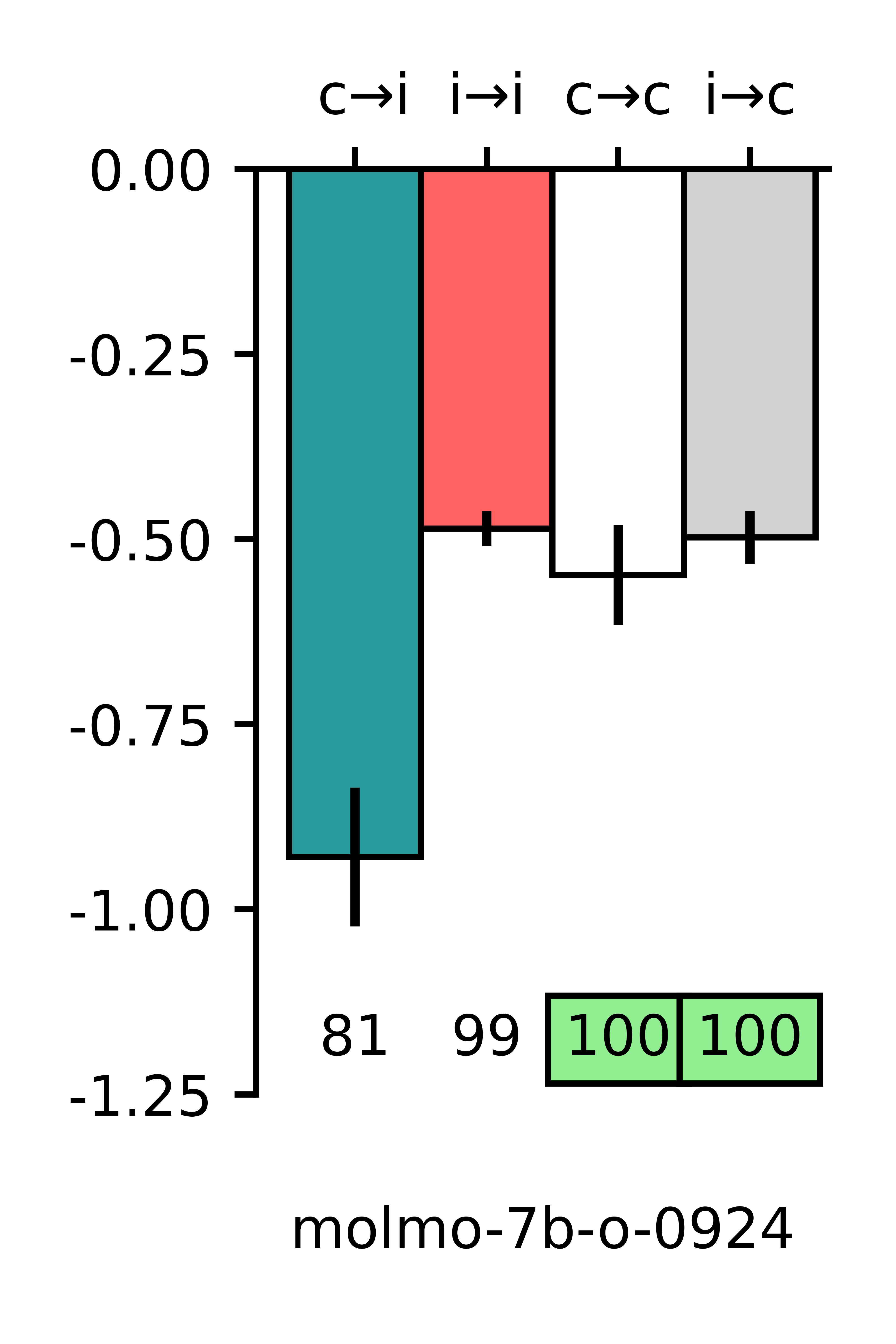}
    \end{subfigure}
    \begin{subfigure}[b]{0.161\textwidth}
        \centering
        \includegraphics[width=\textwidth]{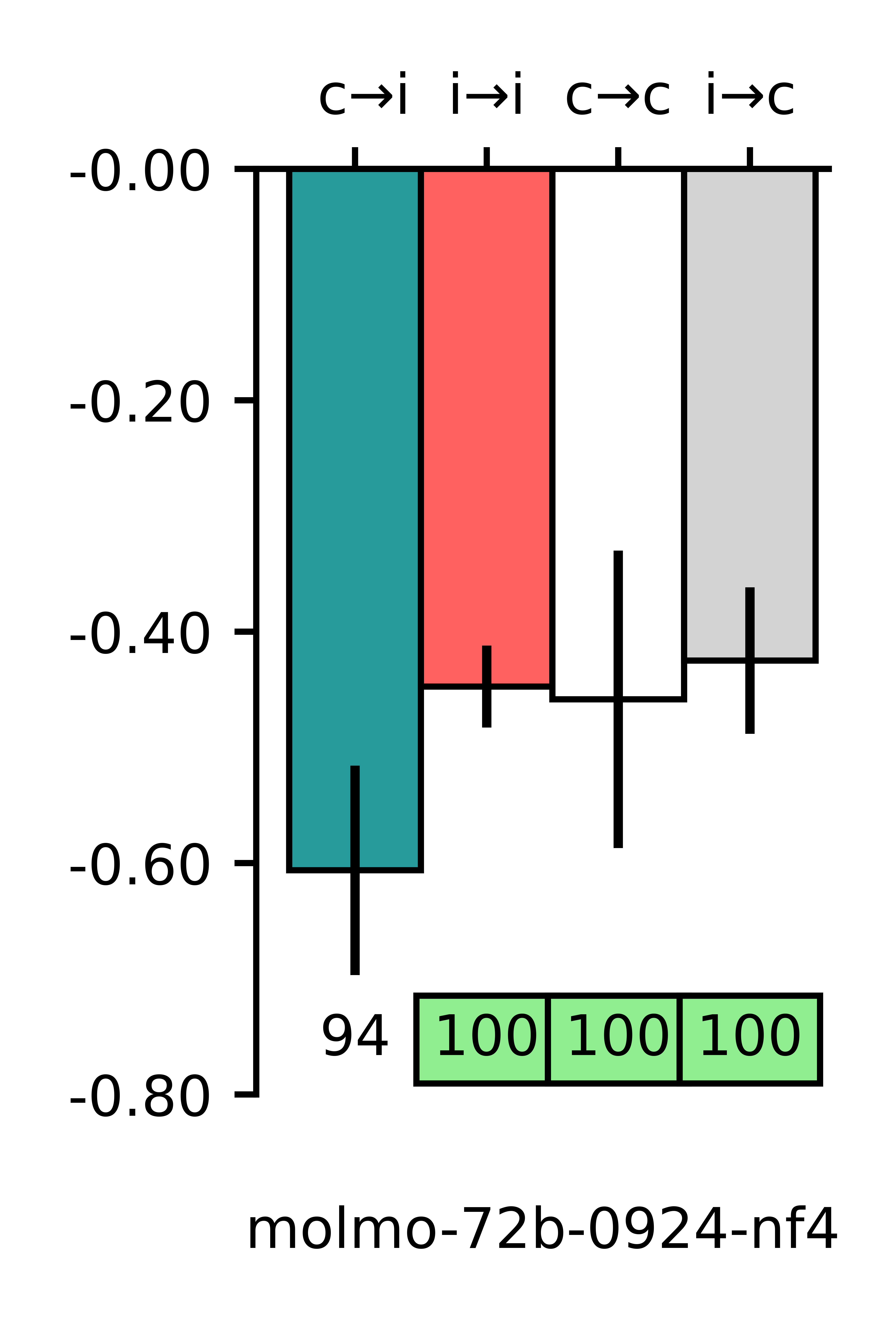}
    \end{subfigure}
    \begin{subfigure}[b]{0.161\textwidth}
        \centering
        \includegraphics[width=\textwidth]{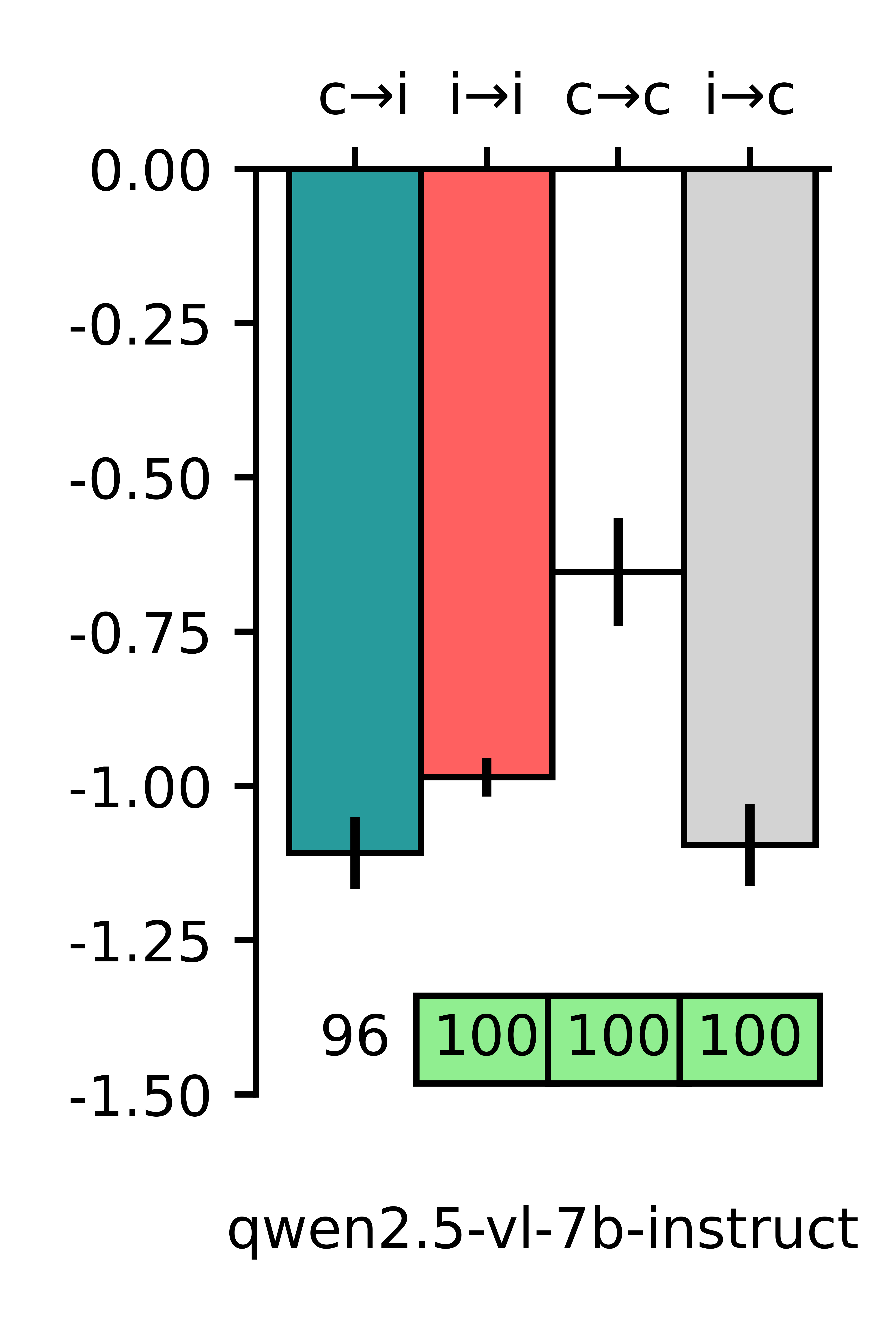}
    \end{subfigure}
    \begin{subfigure}[b]{0.161\textwidth}
        \centering
        \includegraphics[width=\textwidth]{figures/qwen2.5-vl-32b-instruct_top_bottom_logprobs.png}
    \end{subfigure}
    
    \caption{Average log probabilities assigned to correct second color tokens across conditions under top-down word arrangement. 12 of 13 models tested show higher values for II (incongruent following incongruent) than CI (incongruent following congruent). Condition accuracy shown below each bar.}
    \label{fig:model_logprobs_top_bottom}
\end{figure}

\begin{wrapfigure}{r}{0.35\textwidth}
  \centering
  \begin{subfigure}{0.48\linewidth}
    \centering
    \includegraphics[width=\textwidth]{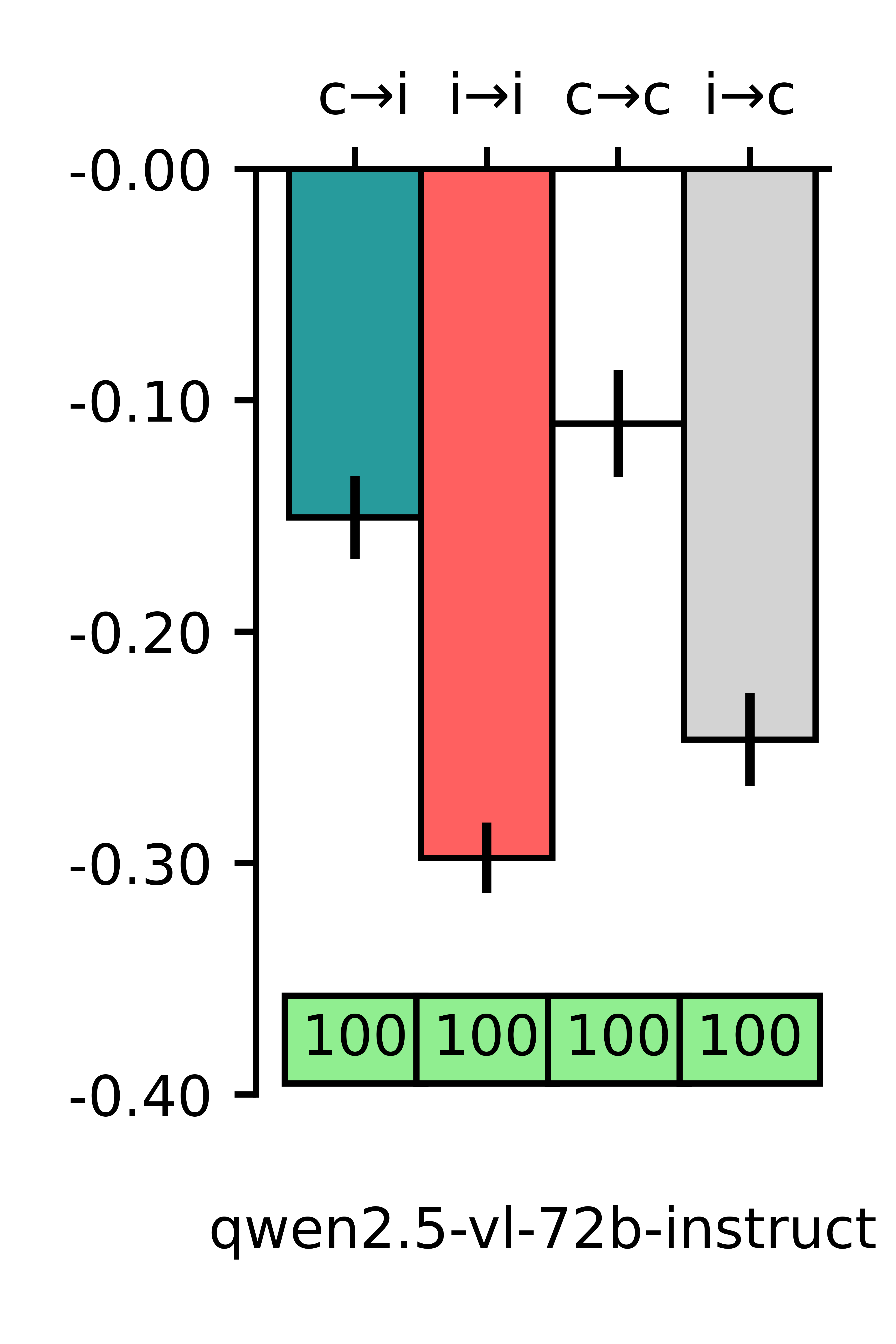}
  \end{subfigure}
  \hfill
  \begin{subfigure}{0.48\linewidth}
    \centering
    \includegraphics[width=\textwidth]{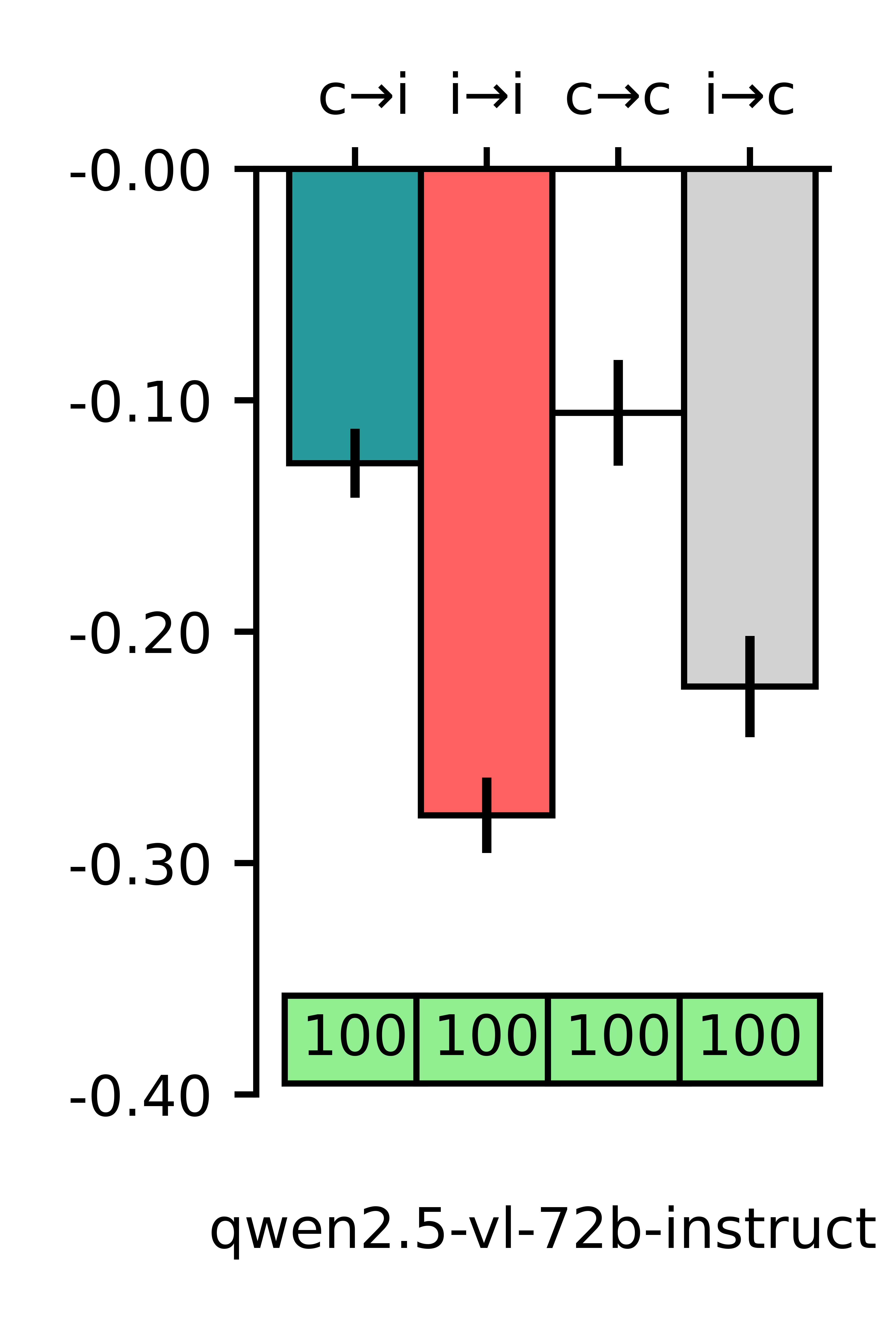}
  \end{subfigure}
  \caption{Qwen2.5 VL 72B Instruct shows reverse conflict adaptation.}
  \vspace{-20pt}
  \label{fig:qwen2.5-vl-72b-instruct_logprobs}
\end{wrapfigure}

To assess conflict adaptation, we compare the average performance on an incongruent trial following a congruent (CI) versus incongruent trial (II).
Figures~\ref{fig:model_logprobs_left_right} and~\ref{fig:model_logprobs_top_bottom} show results across all models for left-right and top-down arrangements, respectively; spatial arrangement did not meaningfully affect the observed pattern.
12 of 13 models demonstrate behavior consistent with conflict adaptation: probabilities for correct second tokens are higher under II compared to CI.
The sole exception is Qwen2.5 VL 72B Instruct, which shows the opposite pattern (Figure~\ref{fig:qwen2.5-vl-72b-instruct_logprobs}).
Given that this model and Qwen2.5 VL 32B Instruct, a model half its size from the same family, both achieve 100\% accuracy across all conditions, we suspect that the task has become trivially easy especially for the 72B model, creating a ceiling effect.

\subsection{Task-Relevant Features}
To understand how VLMs implement conflict processing, we leverage transcoders~\cite{dunefsky2024transcoders}, a variant of SAEs~\cite{bricken2022monosemanticity,cunningham2023sparse}, which seek to decompose dense, polysemantic neural activations into sparse, monosemantic features that respond to specific patterns.
We focus our analysis on InternVL 3.5 4B (pretrained), whose language model component, Qwen 3 4B~\cite{yang2025qwen3}, has available transcoders~\cite{circuit-tracer}.
While applying base language model transcoders to a VLM with modified weights arguably introduces potential concerns, our causal ablation experiments show that the extracted features remain functionally meaningful and interpretable.

To isolate features of interest, we construct ``summary tensors'' by averaging sparse activations across specific trial types and computing differences.
We then apply coactivation-based grouping introduced by Deng et al.~\cite{deng2025sparse} to these summary tensors.
This method constructs inter-layer feature networks based on temporal coactivation across tokens and extracts connected components, which we term ``supernodes'', from each network.
We validate supernode importance through causal ablation: setting supernode features to zero during forward passes and measuring output distribution changes.

\subsubsection{Color and Text Features}

For the color red, we compute summary tensor \texttt{A(c1 == red, t1 != RED, c2 != red, t2 != RED) - A(c1 != red, t1 != RED, c2 != red, t2 != RED)} where \texttt{A} denotes average feature activations and \texttt{ci} and \texttt{ti} respectively denote the \texttt{i}-th color and word. Similarly, for the text RED, we compute summary tensor \texttt{A(c1 != red, t1 == RED, c2 != red, t2 != RED) - A(c1 != red, t1 != RED, c2 != red, t2 != RED)}.

\begin{figure}[htbp]
    \centering
    \includegraphics[width=1\textwidth]{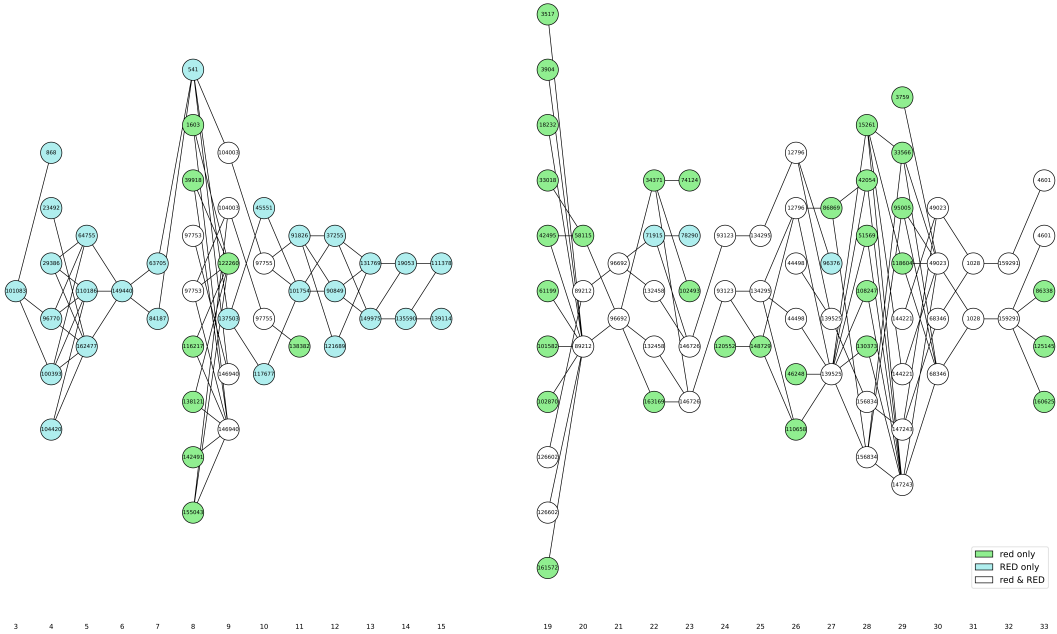}
    \hfill
    \caption{Red (color) supernodes across layers 8-11 and 19-33 and RED (text) supernodes across layers 3-15 and 19-33. Shared features are in white.}
    \label{fig:red_supernodes}
\end{figure}

Figure~\ref{fig:red_supernodes} shows the identified supernodes for the color red and the text RED.
Both text and color form two supernodes each: one in early layers (3-15 for text and 8-11 for color) and one in late layers (19-33 for text and color).
The text and color supernodes partially overlap in both early and late layers, revealing both shared and modality-specific features.
Interestingly, within early layers, the text supernode contains more features than the color supernode; in late layers, the reverse is true.
This mirrors the fact that reading is more automatic than color naming in humans even in the absence of interference.

\subsubsection{Conflict-Modulated Features}

\begin{figure}[htbp]
    \centering
    \includegraphics[width=0.48\textwidth]{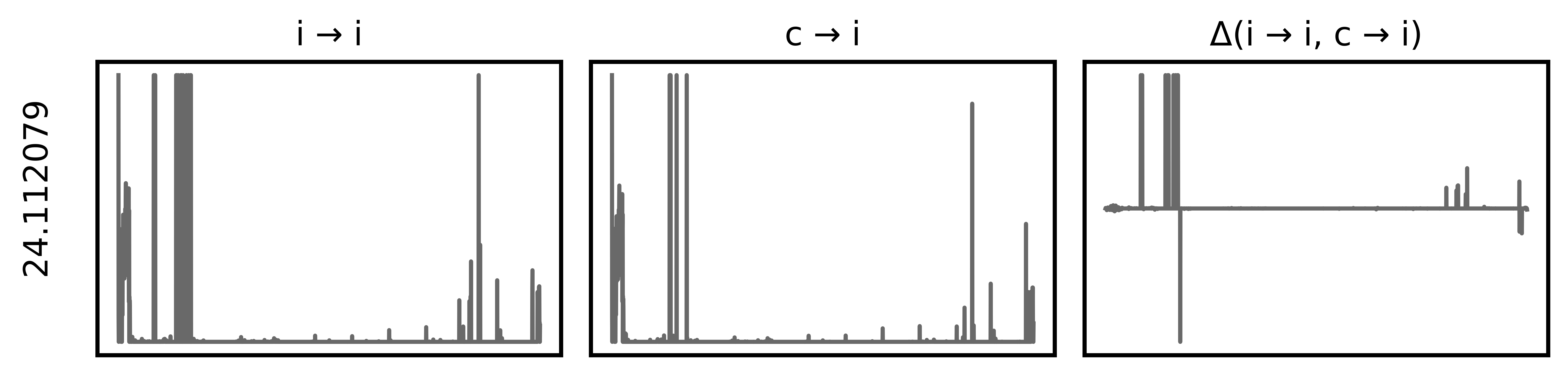}\hfill
    \includegraphics[width=0.48\textwidth]{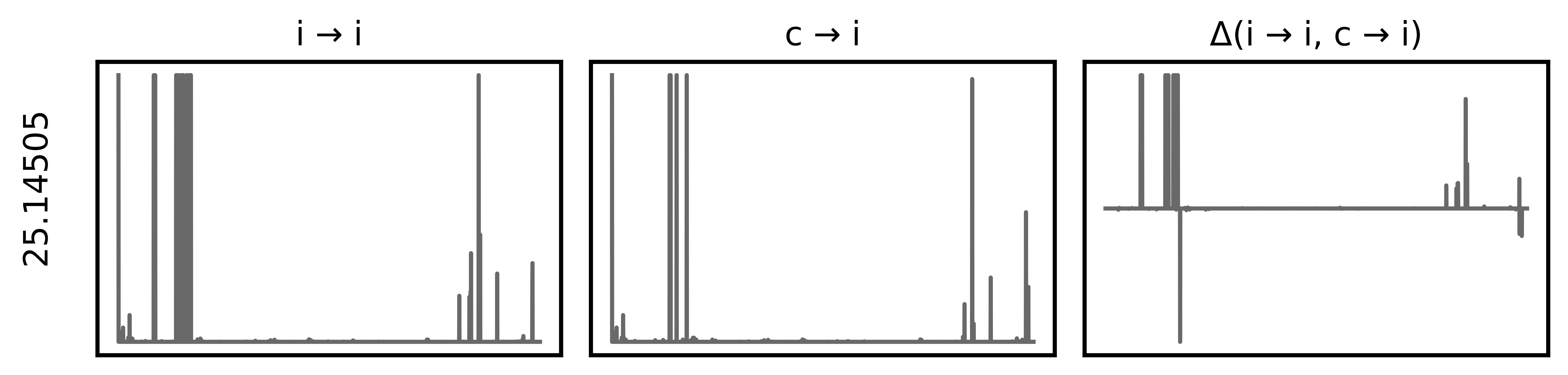}
    
    \vspace{0.5em}
    
    \includegraphics[width=0.48\textwidth]{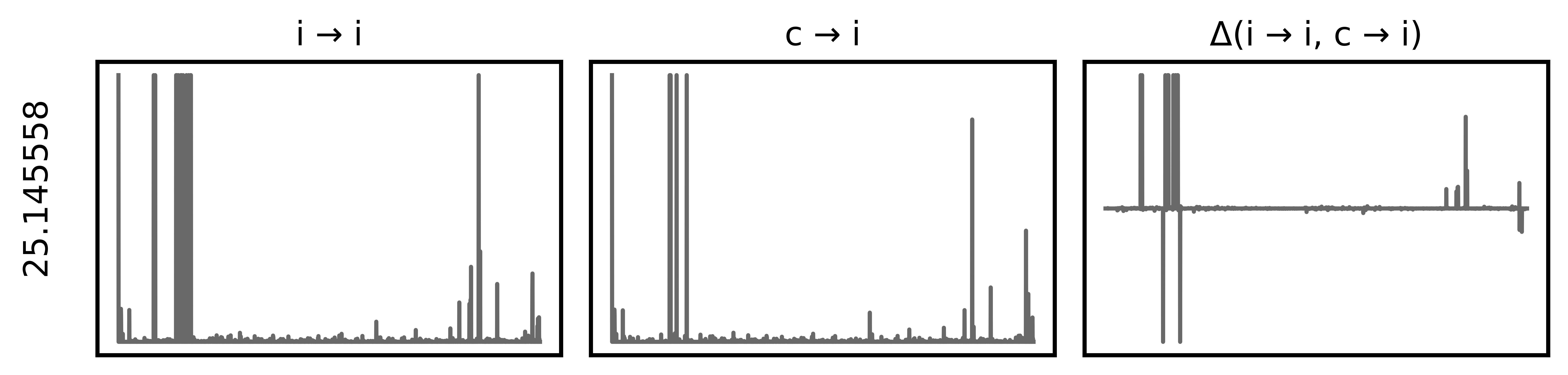}\hfill
    \includegraphics[width=0.48\textwidth]{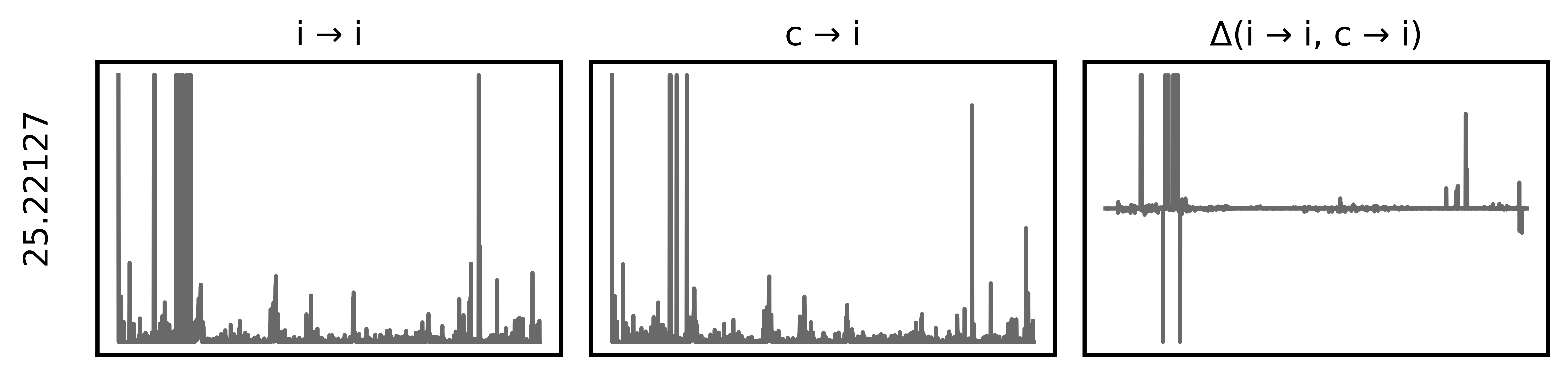}
    
    \vspace{0.5em}
    
    \includegraphics[width=0.48\textwidth]{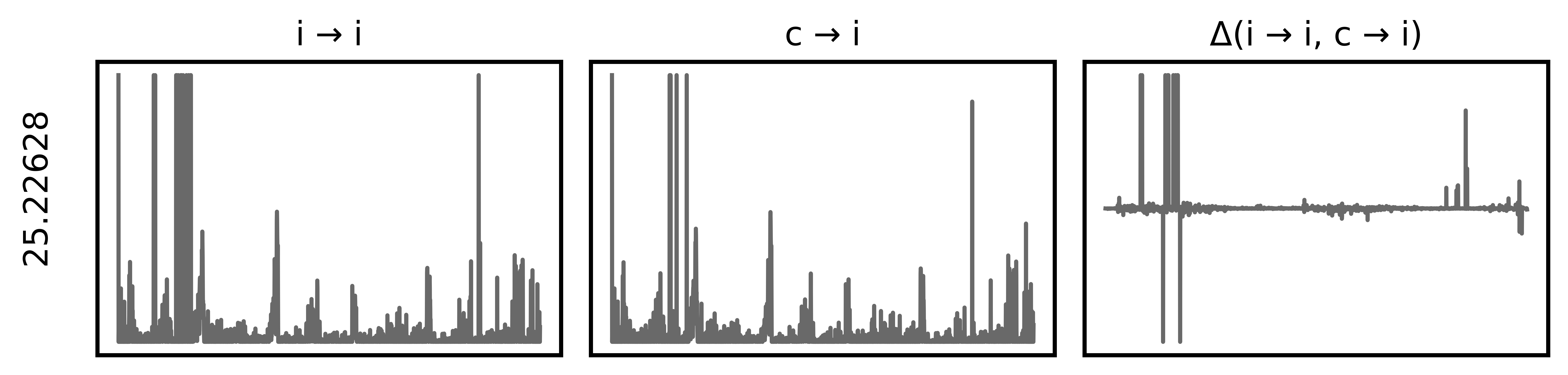}\hfill
    \includegraphics[width=0.48\textwidth]{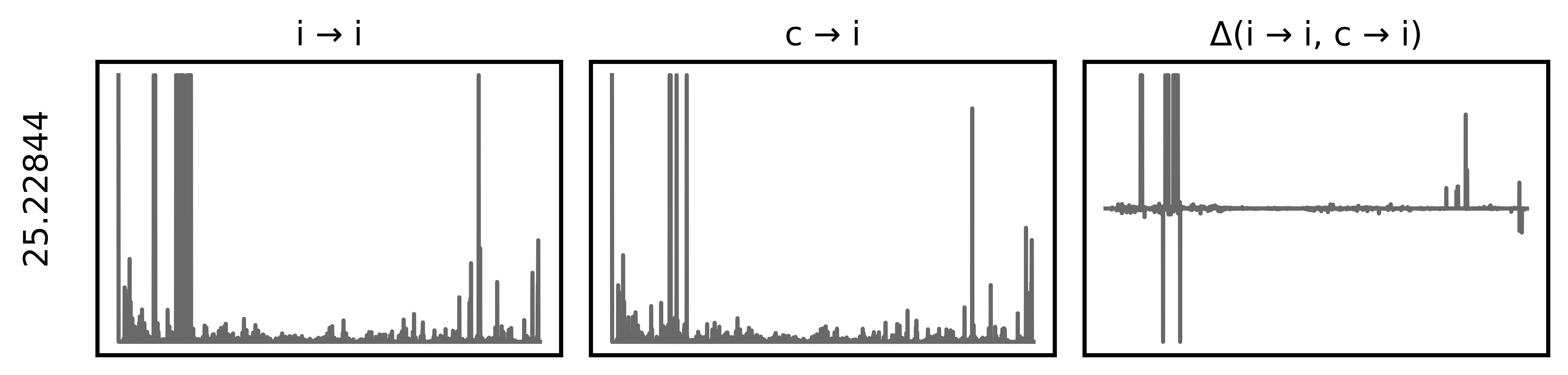}
    
    \caption{Activation patterns across token positions for select features from the conflict-modulated supernode. Columns: II condition, CI condition, and II-CI difference. Histograms truncated at ±80.}
    \label{fig:conflict_activations}
\end{figure}

To identify conflict-related features, we compute summary tensor \texttt{A(c1 != t1, c2 != t2) - A(c1 == t1, c2 != t2)}. Feature coactivation revealed a supernode in layers 24-25 showing more elevated activation under II than CI (Figure~\ref{fig:conflict_activations}).
Unlike color and text features, which are sparse and interpretable (top activating texts largely mention the corresponding colors), the conflict-modulated features are dense—activating frequently across diverse contexts—and lack clear interpretable descriptions. This suggests they serve general rather than task-specific functions.

\begin{figure}[htbp]
    \centering
    \includegraphics[width=0.7\textwidth]{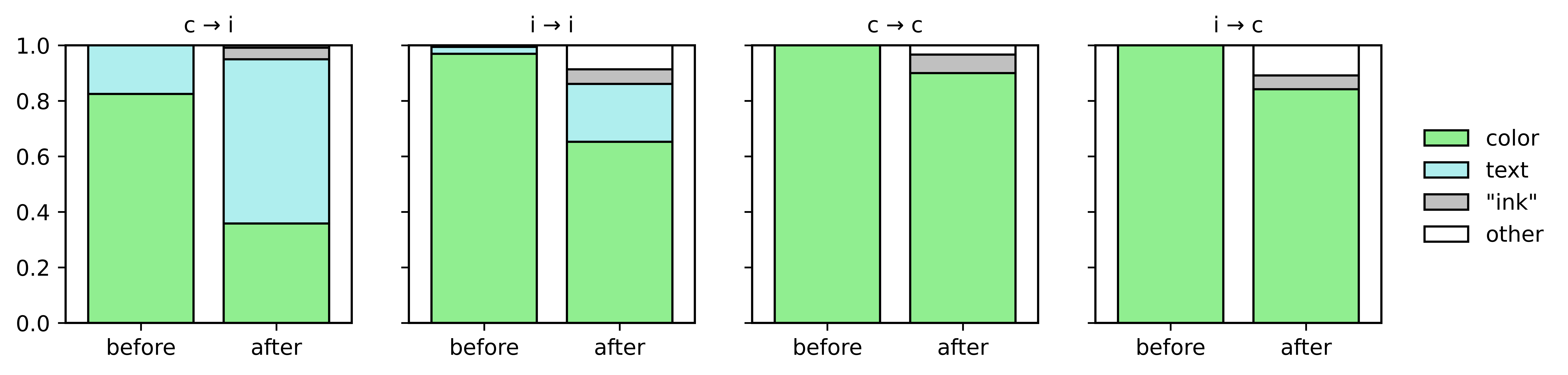}
    \hfill
    \caption{Second-trial output distributions before and after ablating the conflict-modulated supernode. Ablation increases Stroop errors 3.38- and 8.33-fold under CI and II conditions, respectively, with minimal effect on congruent trials (CC, IC).}
    \label{fig:ablation}
\end{figure}

Ablating this supernode increases Stroop errors substantially: 3.38-fold on CI trials (17.5\% to 59.2\%) and 8.33-fold on II trials (2.5\% to 20.8\%), with minimal effect on congruent trials (Figure~\ref{fig:ablation}).
Notably, post-ablation, the model sometimes repeatedly outputs the word ``ink'' in place of the second color, a failure mode entirely absent pre-ablation.
\section{Limitations}\label{sec:discussion}

Task comprehension presents a potential confound in cognitive evaluations of language models~\cite{hu2024auxiliary, hu-lewis-2025-language}. Although we provide thorough instructions and all models demonstrate baseline accuracy, we cannot rule out that conflict may help models better understand task requirements. What appears as conflict adaptation may therefore partly reflect conflict clarifying what the task demands.
\section{Acknowledgements}\label{sec:acknowledgements}

The author would like to thank Sebastian Musslick, Chandra Sripada, Xida Ren, Ruixuan Deng, Richard L. Lewis, Daniel Weissman, Taraz Lee, Shane Storks, Aalok Sathe, and Etha Hua for helpful feedback and discussions.

\bibliographystyle{plain}
\bibliography{references}


\end{document}